\documentclass[journal,compsoc]{IEEEtran}
\pdfoutput=1

\usepackage{pdfsync}

\usepackage[textwidth=1.3cm,textsize=tiny]{todonotes}
\usepackage{graphicx}

\usepackage[cmex10]{amsmath}
\usepackage{amssymb,amsthm}

\usepackage{algorithm}
\usepackage{algorithmic}
\usepackage[tight,normalsize,sf,SF]{subfigure}

\usepackage[caption=false,font=normalsize,labelfont=sf,textfont=sf]{subfig}
\usepackage{url}

\usepackage[utf8]{inputenc}
\usepackage[T1]{fontenc}
\usepackage{amssymb}

\def\bs{\boldsymbol}
\def\cl{\mathcal}

\def\L1{\ell_1}
\def\L{\bs{\cl L}}
\def\R{\mathbb R}
\def\B{\mathbb B}

\def\p{{\bs p}}
\def\b{{\bs b}}
\def\x{{\bs x}}
\def\y{{\bs y}}

\def\LD{{\mathcal L}}
\def\B{{\mathcal B}}
\def\S{{\mathcal S}}

\def\G{\bs {\mathcal G}}

\def\Wave{{\bs W}}
\def\prox{\operatorname{prox}}
\def\proj{\operatorname{proj}}
\def\sign{\operatorname{sign}}
\newcommand{\inv}[1]{\frac{1}{#1}}
\newcommand{\tinv}[1]{{\textstyle\frac{1}{#1}}}
\newcommand{\Id}{{\rm \bf Id}}
\DeclareMathOperator*{\argmin}{\arg\min}

\def\ie{\textit{i.e.},~}
\def\eg{e.g.,~}
\def\st{{\rm s.t.}~}
\newcommand{\scp}[2]{\langle #1, #2 \rangle}

\usepackage{cite}
\usepackage[sort&compress, numbers]{natbib}

\begin{document}
%
%
\title{From Bits to Images:\\ Inversion of Local Binary Descriptors}
%
%
%

\author{Emmanuel~d'Angelo, Laurent~Jacques,  Alexandre~Alahi and~Pierre~Vandergheynst
\IEEEcompsocitemizethanks{\IEEEcompsocthanksitem E. d'Angelo and P. Vandergheynst are with the Signal Processing Labs (LTS2), Ecole Polytechnique Fédérale de Lausanne, 1015 Lausanne, Switzerland. E-mail: \{emmanuel.dangelo,pierre.vandergheynst\}@epfl.ch
\IEEEcompsocthanksitem Alexandre Alahi is with the Stanford Vision Lab, Stanford University, CA 94305-9020, USA. E-mail: alahi@stanford.edu
\IEEEcompsocthanksitem Laurent Jacques is with the ICTEAM institute, ELEN Department, Universit\'e catholique de Louvain (UCL), 1348 Louvain-la-Neuve, Belgium. E-mail: laurent.jacques@uclouvain.be}}

\IEEEcompsoctitleabstractindextext{%
\begin{abstract}
Local Binary Descriptors are becoming more and more popular for image matching tasks, especially when going mobile.
While they are extensively studied in this context, their ability to carry enough information in order to infer the original image is seldom addressed.
In this work, we leverage an inverse problem approach to show that it is possible to directly reconstruct the image content from Local Binary Descriptors.
This process relies on very broad assumptions besides the knowledge of the pattern of the descriptor at hand.
This generalizes previous results that required either a prior learning database or non-binarized features.
Furthermore, our reconstruction scheme reveals differences in the way different Local Binary Descriptors capture and encode image information.
Hence, the potential applications of our work are  multiple, ranging from privacy issues caused by eavesdropping image keypoints streamed by mobile devices to the design of better descriptors through the visualization and the analysis of their geometric content.
\end{abstract}

}

\maketitle

\IEEEpeerreviewmaketitle

\section{Introduction}

\IEEEPARstart{H}ow much, and what type of information is encoded in a keypoint descriptor ?
Surprisingly, the answer to this question has seldom been addressed directly.
Instead, the performance of keypoint descriptors is studied extensively through several image-based benchmarks following the seminal work of Mikolajczyk and Schmid~\cite{Mikolajczyk:2005tk} using Computer Vision and Pattern Recognition task-oriented metrics.
These stress tests aim at measuring the stability of a given descriptor under geometric and radiometric changes, which is a key to success in matching templates and real world observations.
While precision/recall scores are of primary interest when building object recognition systems, they do not tell much about the intrinsic quality and quantity of information that are embedded in the descriptor.
Indeed, these benchmarks are informative about the context in which a descriptor performs well or poorly, but not why.
As a consequence, descriptors were mostly developed empirically by benchmarking new ideas against some image matching datasets.

Furthermore, there is a growing trend towards the use of image recognition technologies from mobile handheld devices such as the smartphones combining high quality imaging parts and a powerful computing platform.
Application examples include image search and landmark recognition~\cite{GoogleGoggles} or augmented media and advertisement~\cite{kooaba}.
To reduce the amount of data exchanged between the mobile and the online knowledge database, it is tempting to use the terminal to extract image features and send only these features over the network.
This data is obviously sensitive since it encodes what the user is viewing.
Hence it is legitimate to wonder if its interception could lead to a privacy breach.

Recently, an inspirational paper \cite{Weinzaepfel:2011jh} showed that ubiquitous interest points such as SIFT~\cite{Lowe:2004uq} and SURF~\cite{Bay:2006ug} suffice to reconstruct plausible source images.
This method is based on an image patch database indexed by their SIFT descriptors and then proceeds by successive queries, replacing each input descriptor by the corresponding patch retrieved in the learning~set.
Although it produces good image reconstruction results, it eventually tells us little about the information embedded in the descriptor: retrieving an image patch from a query descriptor leverages the matching capabilities of SIFT which are now well established by numerous benchmarks and were actually key for its wide adoption.

In this paper, we propose instead two algorithms that aim at reconstructing image patches from local descriptors \emph{without} any external information and with very little additional constraints.
We consider descriptors made of local image intensity differences, which are increasingly popular in the Computer Vision community, for they are not very demanding in computational power and hence well suited for embedded applications.
The first algorithm that we describe works on \emph{real-valued} difference descriptors, and addresses the reconstruction process as a regularized deconvolution problem.
The second algorithm leverages some recent results from 1-bit Compressive Sensing (CS) \cite{Baraniuk:2011wk} to reconstruct image parts from \emph{binarized} difference descriptors, and hence is of great practical interest because these descriptors are usually available as bitstrings  rather than as real-valued vectors.

\subsubsection*{Contributions}

The contributions of this paper are twofold. 
First, we extend the seminal work of \cite{Weinzaepfel:2011jh} by showing that an inverse problem approach suffices to invert a local image patch descriptor provided that the descriptor is a local difference operator, thus avoiding the need to build an external database beforehand.
Second, we present the first 1-bit reconstruction algorithm with practical applications: even Compressive Sensing (CS) related developments are still focused on theoretical issues tested on toy signals.

An earlier version of this work appeared in \cite{dAngelo:2012tl}.
It was however limited to real-valued descriptors, hence we greatly extend it by proposing an algorithm for 1-bit measurements.
We also replace the Total Variation constraint of  \cite{dAngelo:2012tl} by the sparsity of wavelet analysis coefficients in order to have two reconstruction algorithms (for real and binarized descriptors) that optimize over similar quantities in the current paper, indeed easing the reading.
Furthermore, we had to drop the detailed derivation of the real-valued algorithm therein for brevity concerns, and take advantage of the current paper to make the technical steps more explicit.


\subsubsection*{Notations}

In this paper, we make extensive use of the following notations. Matrices and vectors are denoted by bold letters or symbols (\eg $\bs \Phi$, $\bs x$)  while light letters are associated to scalar values (\eg scalar functions, vector components or dimensions).
The scalar product between two $N$-length vectors $\bs x$ and $\bs y$ is written $\langle \x,\y \rangle = \sum_{i=1}^N x_i y_i$, while their Hadamard product $\x \odot \y$ is such that $(\x \odot \y)_i =  x_i y_i$ for $1\leq i\leq N$.
Since we work only with real matrices, the adjoint of a matrix $\bs A$ is $\bs A^*=\bs A^T$.
The vector of ones is written $\bs 1=(1,\cdots,1)^T$ and the identity matrix is denoted $\Id$.
    
Most of the time, we will ``vectorize'' 2-D images, \ie an image or a patch image $\bs x$ of dimension $N_1\times N_2$ is represented as a $N$-dimensional vector $\bs x\in \R^N$ with $N=N_1N_2$.
This allows us to represent any linear operation on $\bs x$ as a simple matrix-vector multiplication.
One important linear operator is the 2-D wavelet analysis operator $\Wave$ with $\Wave^T$ the corresponding synthesis operator.
For $\bs x,\bs y \in \R^N$, $\Wave \bs x$ is then a vector of wavelet coefficients and $\Wave^T \bs y$ a patch with the same size as~$\bs x$. 

We denote by $\|\bs x\|_p=(\sum_i |x_i|^p)^{1/p}$ with $p\geq 1$ the $\ell_p$-norm of $\bs x\in \R^N$, reserving the notation $\|\cdot\|$ for $p=2$ and with $\|\bs x\|_\infty = \max_i |x_i|$.
The $\ell_0$ ``norm'' of $\bs x$ is $\|\bs x\|_0 = \#\{i: x_i \neq 0\}$. Correspondingly, for $1\leq p\leq +\infty$, a $\ell_p$-ball of radius $\lambda$ is the set $B_p(\lambda) = \{\bs x\in\R^N: \|\bs x\|_p \leq \lambda\}$. 

We use also the following functions. 
We denote by $(\bs x)_+$ the non-negativity thresholding function, which is defined componentwise as $(\lambda)_+ = (\lambda + |\lambda|)/2$, and $(\bs x)_- = -(-\bs x)_+$.
The sign function $\sign \lambda$ is equal to 1 if $\lambda >0$ and $-1$ otherwise.

In the context of convex optimization, we denote by $\Gamma^0(\R^N)$ the class of proper, convex and lower-semicontinuous functions of the finite dimensional vector space $\R^N$ to $(-\infty,+\infty]$~\cite{Combettes:2011wd}.
The indicator function $\imath_\S\in \Gamma^0(\R^N)$ of a set $\S$ maps $\imath_\S(\bs x)$ to 0 if $\bs x \in \S$ and to $+\infty$ otherwise.
For any $F\in \Gamma^0(\R^N)$ and $\bs z\in \R^N$, the \emph{Fenchel-Legendre} conjugate function $F^*$ is 
\begin{equation}
	F^* (\bs z) = \max_{\bs x\in\R^N}\ \langle \bs z, \bs x \rangle - F(\bs x),
\label{eq:legendre}
\end{equation}
while, for any $\lambda >0$, its \emph{proximal operator} reads
\begin{equation}
	\prox_{\lambda F} \bs z = \argmin_{\bs x\in\R^N}\ \lambda F(\bs x) + \tinv{2}\| \bs x - \bs z \|^2.
	\label{eq:prox}
\end{equation}
For $F=\imath_{\cl S}$ for some convex set $\cl S\subset \R^N$, the proximal operator of $\prox_{\lambda F}$ simply reduces to the orthogonal projection operator on $\cl S$ denoted by $\proj_{\cl S}$.

\section{Local Binary Descriptors}
\label{sec:lbd}

In this paper, we are interested in reconstructing image patches from binary descriptors obtained by quantization of local image differences, such as BRIEF \cite{Calonder:2010tx} or FREAK \cite{Anonymous:_4iTrXEX}.
Hence, we will refer to these descriptors as Local Binary Descriptors (LBDs) in the sequel.
In a standard Computer Vision and Pattern Recognition application, such as object recognition or image retrieval, an interest point detector such as Harris corners \cite{Harris:1988tk}, SIFT~\cite{Lowe:2004uq} or FAST \cite{Rosten:ws} is first applied on the images to locate interest points.
The regions surrounding these keypoints are then described by a feature vector, thus replacing the raw light intensity values by more meaningful information such as histograms of gradient orientation or Haar-like analysis coefficients.
In the case of LBDs, the feature vectors are made of local binarized differences computed according to the generic process described below.

\subsection{Generic Local Binary Descriptor model}

A LBD of length $M$ describing a given image patch of $\sqrt{N}\times\sqrt{N} = N$ pixels can be computed by iterating $M$ times the following three-step process:
\begin{enumerate}
\item compute the Gaussian average of the patch at two locations $\bs x_i$ and $\bs x'_i$ with variance $\sigma_i$ and $\sigma_i'$ respectively;
\item form the difference between these two measurements;
\item binarize the result by retaining only its sign.
\end{enumerate}

\begin{figure}[!t]
	\centering
        \includegraphics[width=.7\columnwidth]{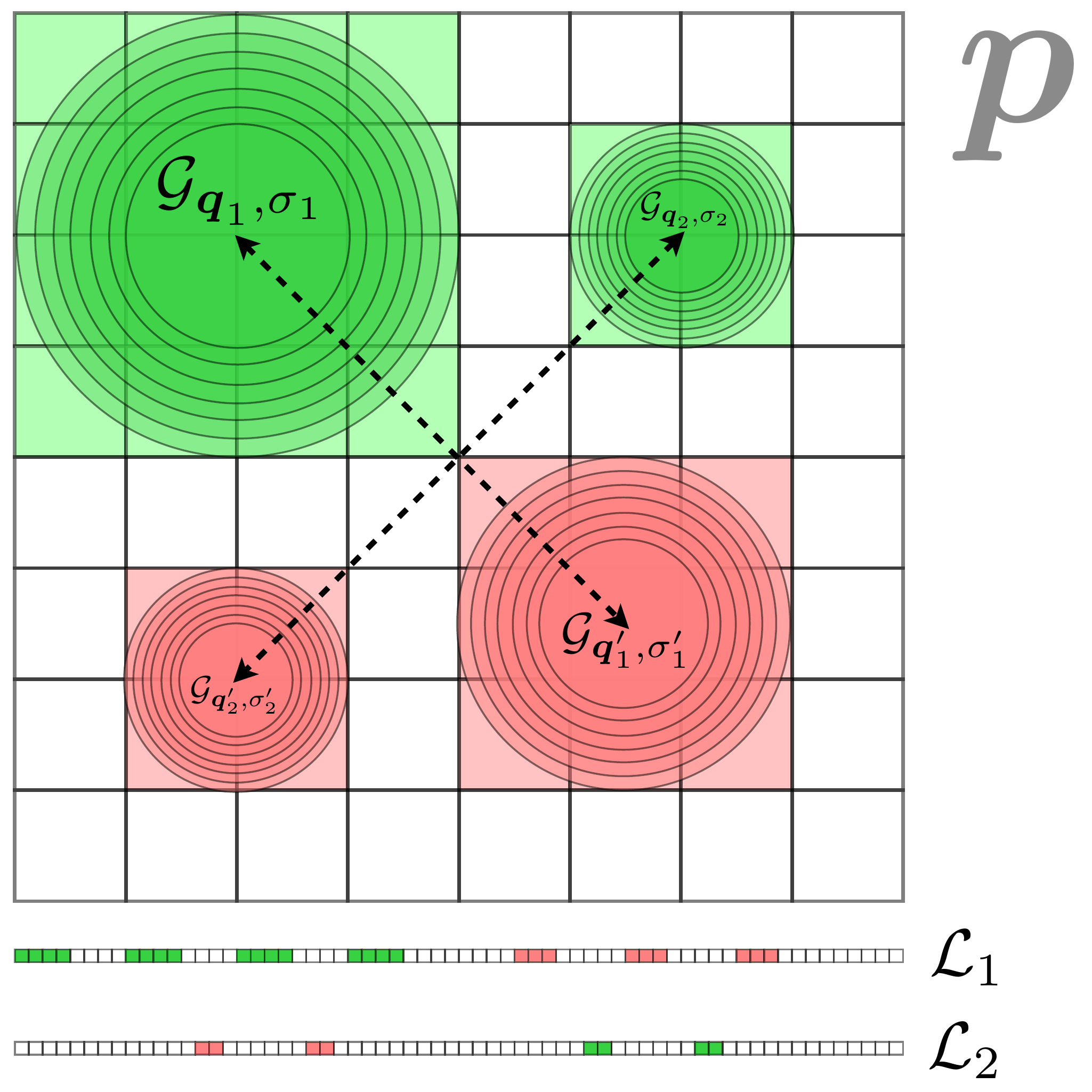}
	\caption{Example of a local descriptor for an $8 \times 8$ pixels patch and the corresponding sensing matrix. %
			Only two measurements of the descriptor are depicted; each one is produced by subtracting the Gaussian mean in the lower (red) area from the corresponding upper (green) one. %
			All the integrals are normalized by their area to have values in~$[0,1]$. In the bottom, the corresponding vectors.}
	\label{fig:lbd}
\end{figure}

Reshaping the input patch as a column vector $\mathbf p \in \R^N$, the first two steps in the above procedure can be merged into the application of a single linear operator $\bs\LD$:
\begin{align}
\bs \LD: 	\R^N & \rightarrow \R^M \notag \\
		\p & \mapsto \big(\langle\G_{\bs q_i, \sigma_i} , \p \rangle - \langle \G_{\bs q_i', \sigma_i'}, \p\rangle\big)_{1\leqslant i \leqslant M}\,, \label{eq:L}
\end{align}
where $\G_{\bs q, \sigma}\in\R^N$ denotes a (vectorized) two-dimensional Gaussian of width~$\sigma$ centered in $\bs q\in\R^2$ (Fig.~\ref{fig:lbd}, top) with $\|\G_{\bs q, \sigma}\|_1=1$.
As illustrated in Fig.~\ref{fig:lbd}-bottom, since $\bs \LD$ is a linear operator, it can be represented by a matrix $\bs{\cl L}\in \R^{M\times N}$ multiplying $\bs p$ and whose each row $\bs {\cl L}_i$ is given by 
\begin{equation}
\bs {\cl L}_i = \G_{\bs q_i, \sigma_i}- \G_{\bs q_i', \sigma_i'},\quad 1\leq i\leq M.
\label{eq:Li}
\end{equation}
We will take advantage of this decomposition interpretation to avoid explicitly writing $\bs \LD$ later on.

The final binary descriptor is obtained by the composition of this sensing matrix with a component-wise quantization operator $\B$ defined~by $
	\B(\x)_i = \sign x_i$, so that, given a patch $\bs p$, the corresponding LBD reads
\begin{equation}
  \bar{\bs p} := \cl B(\bs{\cl L} \bs p)\ \in\ \{-1,+1\}^M.
  \label{eq:final-lbd-def}
\end{equation}
Note that we have chosen this definition of $\B$ to be consistent with the notations of~\cite{Baraniuk:2011wk}.
Implementations of LBDs will of course use the binary space $\{0,1\}^M$ instead, since it fits naturally with the digital representation found in computers.

\begin{figure}[!t]
	\centering
	\subfigure[$\L_1$]{\includegraphics[width=.225\columnwidth]{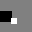}}\hfill
	\subfigure[$\L_2$]{\includegraphics[width=.225\columnwidth]{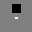}}\hfill
	\subfigure[$\L_3$]{\includegraphics[width=.225\columnwidth]{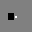}}\hfill
	\subfigure[$\L_4$]{\includegraphics[width=.225\columnwidth]{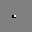}}
	\newline
	\subfigure[$\L_{509}$]{\includegraphics[width=.225\columnwidth]{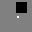}}\hfill
	\subfigure[$\L_{510}$]{\includegraphics[width=.225\columnwidth]{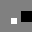}}\hfill
	\subfigure[$\L_{511}$]{\includegraphics[width=.225\columnwidth]{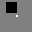}}\hfill
	\subfigure[$\L_{512}$]{\includegraphics[width=.225\columnwidth]{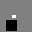}}\hfill
	\caption{The first and last frame vectors for a FREAK descriptor of 512 measurements (2-D representations).
			The white square depicts the positive lobe and the dark square the negative one.
			Each approximated Gaussian is normalized to be of unit $\ell_1$-norm.}
	\label{fig:Li}
\end{figure}

From the description of LBDs, it is clear that they involve only simple arithmetic operations.
Furthermore, the distance between two LBDs is measured using the Hamming distance, which is a simple bitwise exclusive-or (XOR) instruction~\cite{Calonder:2010tx,Anonymous:_4iTrXEX}.
Hence, computation and matching of LBDs can be implemented efficiently, sometimes even using hardware instructions (XOR), allowing their use on mobile platforms where computational power and electric consumption are strong limiting constraints.
Since they also provide good matching performances, LBDs are getting more and more popular over SIFT and SURF: combined with FAST for the keypoint detection, they provide a fast and efficient feature extraction and matching pipe-line, producing compact descriptors that can be streamed over networks.

Typically, a 32-by-32 pixels image patch (1024 bytes in 8 bit grayscale format) can be reduced to a vector of only 256 measurements~\cite{Calonder:2010tx} coded with 256 \emph{bits}.
A typical floating-point descriptor such as SIFT or SURF would require instead 64 float values, \ie 256 \emph{bytes} for the same patch, eight times the LBD size, and the distances would be measured with the $\ell_2$-norm using slower floating-point instructions.

\subsection{LBDs, LBPs, and other integral descriptors}

Unlike \cite{Weinzaepfel:2011jh}, we use LBDs in this work instead of SIFT descriptors.
As we will see in Section~\ref{sec:reconstruction}, it is actually the knowledge of the spatial measurement pattern used by an LBD that allows us to properly define the matrix of the operator $\LD$ in \eqref{eq:L} as a convolution matrix.
SIFT and SURF use histograms of gradient orientation instead, thus losing the precise localization information through an integration step.
Hence, it seems very unlikely that our approach could be extended to these descriptors.
On the other hand however it is possible to reproduce most of the algorithm described in \cite{Weinzaepfel:2011jh} by replacing SIFT with a correctly chosen LBD to index the reference patch database, but this would bring only minor novelty.

Note also that we have coined the descriptors used here as LBDs, which are not the same as the Local Binary Patterns (LBPs) popularized by \cite{Ahonen:2006gr} for face detection.
Although both LBDs and LBPs produce bit string descriptors, LBPs are obtained after binarization of image direction histograms.
As such, LBPs are integral descriptors and suffer from the same lack of spatial awareness as SIFT and SURF.

\subsection{The BRIEF and FREAK descriptors}

Given two LBDs, the differences reside in the pattern used to select the size and the location of the measurement pairs $(\G_{\bs q_i,\sigma_i},\G_{\bs q'_i,\sigma'_i})_{i=1}^M$.
The authors of the pioneering BRIEF~\cite{Calonder:2010tx} chose small Gaussians of fixed width to bring some robustness against image noise, and tested different spatial layouts.
Among these, two random patterns outperformed the others: the first one corresponds to a normal distribution of the measurement points centered in the image patch, and the second one to a uniform distribution.

Working on improving BRIEF, the authors of ORB~\cite{Rublee:ug} introduced a measurement selection process based on their matching performance and retained pairs with the highest selectivity.
On the other hand, BRISK~\cite{Anonymous:frctOhD6} introduced a concentric pattern to distribute the measurements inside the patch but retained only the innermost points for the descriptor, keeping the peripheral ones to estimate the orientation of the keypoint.

\begin{figure}[htbp]
	\centering
		\includegraphics[width=2.5in]{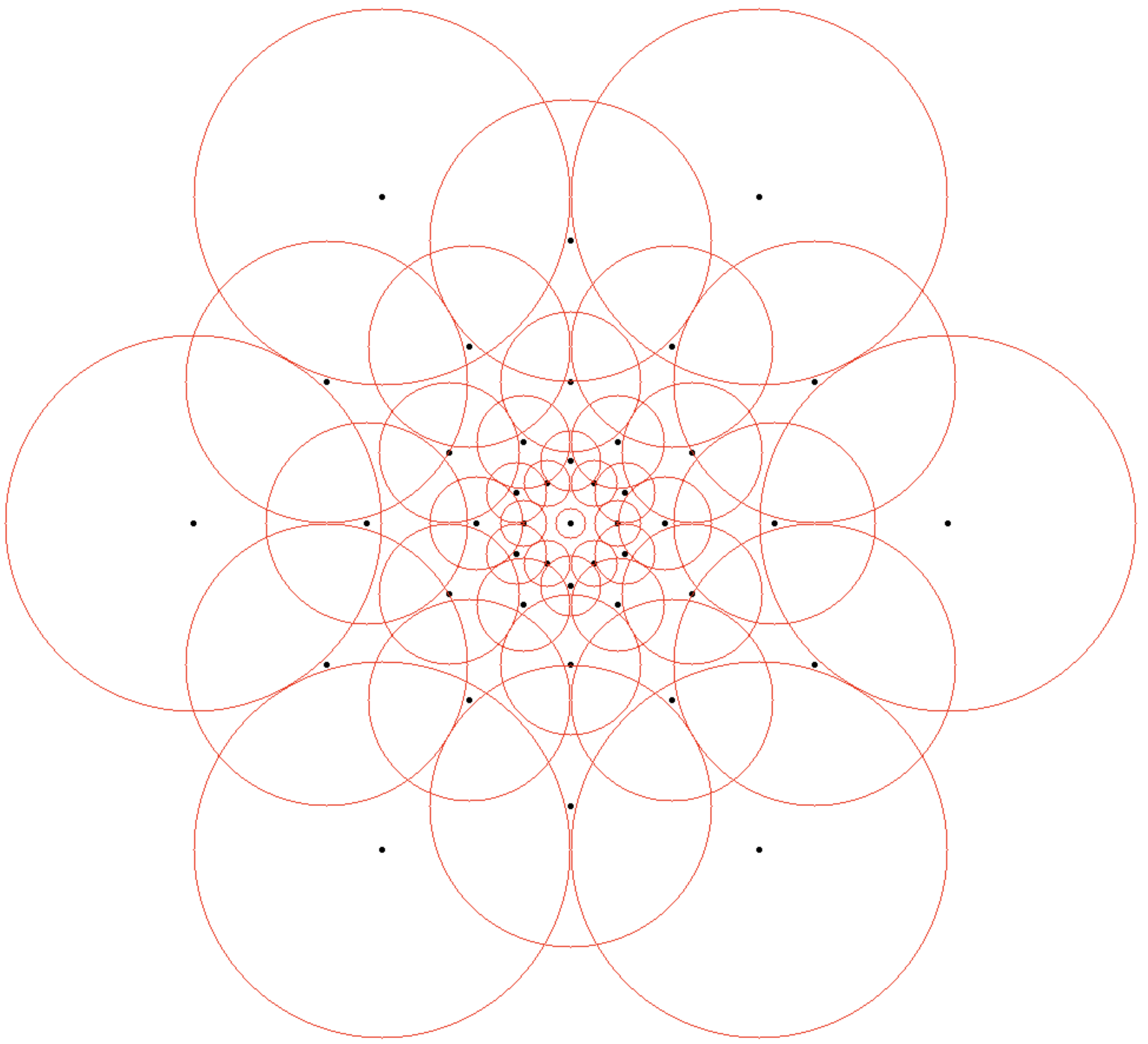}
	\caption{The retinal pattern used by FREAK. %
	The further a point from the center, the wider the averaging area. %
	 Hence, FREAK captures the image variations at a coarser scale on the border of the patch than in its center.}
	\label{fig:freak}
\end{figure}

Eventually, the FREAK descriptor was proposed in~\cite{Anonymous:_4iTrXEX} to leverage the advantages of both approaches: the learning procedure introduced with ORB and the concentric measurement layout of BRISK.
The pattern was modified to resemble the retinal sampling and can be seen in Figure~\ref{fig:freak}.
Note that it allows for a wider overlap than the BRISK pattern.
All the rings were allowed to contribute in the training phase.
Consequently, the FREAK descriptor implicitly captures the  image details at a coarser scale when going away from the center of the patch.

\section{Reconstruction as an inverse problem}
\label{sec:reconstruction}

In this work, our goal is to demonstrate that the knowledge of the particular measurement layout of an LBD is sufficient to infer the original image patch \emph{without any external information}, using only an inverse problem approach.
Typically, a $32 \times 32$ pixels patch (1024 values) will be represented by a descriptor with 512 components.
Hence, the reconstruction task is ill-posed: even without binarization of the features, there are half less measurements than unknowns.
Assuming that this feature vector is represented with floating-point values, the binarization will then divide by an additional factor of 32 (the standard size of a float in bits) the amount of available information!
Classically, to make this problem tractable we introduce a regularization constraint that should be highly generic since we do not know a priori the type of image that we need to reconstruct.
Thus, the sparsity of the reconstructed patch in some wavelet frame appeared as a natural choice: it only imposes that a patch should have few nonzero coefficients when analyzed in this wavelet frame, which is quite general.

\subsection{Real-valued descriptor reconstruction with convex optimization}

Ignoring first the quantization operator by replacing $\B$ by the identity function, we choose the $\ell_1$-norm to penalize the error in the data term and the $\ell_1$-norm of the wavelet coefficients as a sparsity promoting regularizer.
The $\ell_1$-norm is more robust than the usual $\ell_2$-norm to the actual value of the error and it is more connected with its sign.
Hence, it is hopefully a better choice when dealing with binarized descriptors.
The problem of reconstructing an image patch $\hat{\p}\in\R^N$ given an observed binary descriptor $\bar{\bs p} \in \R^M$ then reads:
\begin{equation}
\hat{\p} = \argmin_{\bs x \in \R^N}
\lambda \| \bs\LD \bs x - \bar{\bs p}\|_1
+ 
\| \Wave \bs x \|_1 + \imath_{\mathcal S}(\bs x),
\label{eq:real_inverse_problem}
\end{equation}
which is a sparse $\ell_1$ deconvolution problem \cite{Sidky:2011ud}.
In Eq.~\eqref{eq:real_inverse_problem}, $\|\bs\LD \bs x - \bar{\bs p}\|_1$ is the \emph{data term} that ties the solution to the observation~$\bar{\bs p}$, $\|\Wave x\|_1$ is the regularizer that constrains the patch candidate~$\bs x$ to have a sparse representation, and $\imath_\S(.)$ is the indicator function of the \emph{validity domain} of $\bs x$ that we will make explicit later.

While the objective function in \eqref{eq:real_inverse_problem} is convex, it is not differentiable since the $\ell_1$-norm has singular points on the axes of $\R^N$. 
Hence, we chose the \emph{primal-dual} algorithm presented in \cite{Chambolle:2010hk} to solve this minimization problem. 
Instead of using derivatives of the objective which may not exist, it relies on \emph{proximal calculus} and proceeds by alternate minimizations on the primal and dual unknowns.

The generic version of this algorithm aims at solving minimization problems of the form:
\begin{equation}
	\bs{\hat{x}} = 
	\argmin_{\bs x\in\R^N} F(\bs K\bs x) + G(\bs x),
	\label{eq:primal}
\end{equation}
where $\bs x\in\R^N$ is the \emph{primal} variable, $\bs K\in\R^{D\times N}$ is a linear operator, and $F\in\Gamma^0(\R^D)$ and $G\in\Gamma^0(\R^N)$ are convex (possibly non-smooth) functions. 
The algorithm proceeds by restating (\ref{eq:primal}) as a \emph{primal-dual} saddle-point problem on both the \emph{primal} $\bs x$ and its \emph{dual} variable $\bs u$:
\begin{equation}
	\underset{\bs x}{\min~} \underset{\bs u}{\max~} \langle \bs K\bs x, \bs u\rangle + G(\bs x) -F^* (\bs u).
	\label{eq:canonical_pd}
\end{equation}

For the problem at hand, we start by decoupling the data term and the sparsity constraint in Eq.~\eqref{eq:real_inverse_problem} by introducing the auxiliary unknowns $\bs y, \bs z \in \R^N$ such that:
\begin{equation}
\hat{\p} = \argmin_{\bs y,z \in \R^N}
\lambda \| \bs\LD \bs y - \bar{\bs p}\|_1
+ 
\| \Wave \bs z \|_1 + \imath_{\mathcal S}(\bs y) + \imath_{\{\bs 0\}}(\bs y - \bs z).
\end{equation}
The term $\imath_{\{\bs 0\}}(\bs y - \bs z)$ guarantees that the optimization occurs on the bisector plane $\bs y = \bs z$.    
Then, defining $\bs K=\big(\begin{smallmatrix}\bs{\cl L}&0\\0& \Wave \end{smallmatrix}\big) \in \R^{(M+N)\times 2N}$, we can perform our optimization~\eqref{eq:real_inverse_problem} in the product space $\bs x = (\bs y^T, \bs z^T)^T\in \R^{2N}$~\cite{raguet-gfb,Gonzalez:2012}. 
In this case, \eqref{eq:real_inverse_problem} can be written as (\ref{eq:primal}) by setting $F(\bs K\bs x) = F_1(\bs{\cl L} \bs y) + F_2(\Wave \bs z)$, where:
\begin{align}
F_1(\cdot) &=\lambda\|\cdot - \bar{\bs p}\|_1,\\
F_2(\cdot) &=\|\cdot\|_1,\\
G\big(\begin{smallmatrix}\bs y\\ \bs z\end{smallmatrix}\big) &= \imath_{\cl S}(\bs y) + \imath_{\{\bs 0\}}(\bs y - \bs z).
\end{align}


Introducing $\bs r\in\R^M$ and $\bs s\in\R^{N}$ the dual counterparts of $\bs y$ and $\bs z$ respectively, and taking the Fenchel-Legendre transform of $F$ yields the desired primal-dual formulation of \eqref{eq:real_inverse_problem}:
\begin{equation}
	\min_{\bs y,\bs z}\max_{\bs r,\bs s}\ \langle \bs{\cl L}\bs y , \bs r  \rangle + \langle \Wave \bs z,\bs s \rangle + G\big(\begin{smallmatrix}\bs y\\ \bs z\end{smallmatrix}\big) - F_1^* (\bs r) - F_2^*(\bs s).
	\label{eq:pd_functional}
\end{equation}
An explicit formulation of $F_1^*(\bs r)$ can be obtained by noting that, for $\varphi(\bs r)=\varphi'(\bs r-\bs u)$, $\varphi^*(\bs r)=\varphi'^*(\bs r)+\scp{\bs r}{\bs u}$, and that the conjugate of the $\ell_1$-norm is $\imath_{B_{\infty}(1)}$ \cite{Combettes:2011wd}.
$F_2^*$ is derived in~\cite{Chambolle:2010hk}, eventually yielding:
\begin{align}
F_1^*(\bs r) &= \imath_{B_\infty(\lambda)} ( \bs r ) + \scp{\bs r}{\bar{\bs p}}, \\
F_2^*(\bs s) &= \imath_{B_\infty(1)} ( \bs s ).
\end{align} 
 
The algorithm presented in \cite{Chambolle:2010hk} requires explicit solutions for the proximal mappings of $F_1^*$, $F_2^*$ and $G$.
The first two are easily computed pointwise \cite{Combettes:2011wd,Chambolle:2010hk} as:
\begin{align}
	(\prox_{\sigma F_1^*} \bs r)_i&= \sign(r_i - \sigma \bar p_i) \cdot  \min(\lambda, |r_i - \sigma \bar p_i|), \label{eq:proxf1star}\\
	(\prox_{\sigma F_2^*} \bs s)_i&= \sign(s_i) \cdot  \min(1, |s_i|). \label{eq:proxf2star}
\end{align}

The function $G$ is formed by the indicator of the set $\S$ and the indicator of the bisector plane $\{\big(\begin{smallmatrix}\bs y\\ \bs z\end{smallmatrix}\big)\in\R^{2N}:\bs y=\bs z\}$. 
An easy computation provides \cite{Gonzalez:2012}:
\begin{equation}
\prox_{\sigma G} \big(\begin{smallmatrix}\bs y\\ \bs z \end{smallmatrix}\big) = \begin{pmatrix}\proj_{\cl S} \inv{2}(\bs y + \bs z)\\[1mm] \proj_{\cl S} \inv{2}(\bs y + \bs z)\end{pmatrix}.
\end{equation}
Thus, its proximal mapping does not depend on any parameter. 

Let us now precise our \emph{validity domain} $\cl S$.
It is defined in order to remove ambiguities in the definition of the program (\ref{eq:real_inverse_problem}) that could lead to a non-uniqueness of the solution.
They are due to the differential nature of $\bs{\cl L}$, \ie the descriptor of any constant patch is zero.
This involves both that $\bar{\bs p}$ does not include any information about the average of the initial patch $\bs p$, and the average of $\bs x$ in (\ref{eq:real_inverse_problem}) cannot be determined by the optimization.

This problem is removed by defining $\cl S$ as the intersection of two convex sets $\cl S_1$ and $\cl S_2$. 
The first set $\cl S_1$ arbitrarily constrains the minimization domain to stay in the set of patches whose pixel dynamic lies in $[0,h_{\rm pix}]$, \ie $\cl S_1=\{\bs x\in\R^N: 0\leq x_i\leq h_{\rm pix}\}$. 
In our experiments, we simply consider pixels with real values in $[0,1]$ and consequently fix $h_{\rm pix}=1$. 
The second domain $\cl S_2$ is associated to the space of patches whose pixel mean is equal to $0.5$, \ie $\cl S_2=\{\bs x\in\R^N: \tinv{N}\,\sum_i x_i = 0.5\}$.   

This gives us a first set $\S_1$ whose proximal mapping $\proj_{\S_1}$ is a simple clipping of the values in $[0,1]$, while $\cl S_2$ is an hyperplane in $\R^N$ whose corresponding proximal mapping $\proj_{\S_2}$ is the projection onto the simplex of $\R^N$ of vectors with mean $0.5$.
This projection can be solved efficiently using~\cite{Michelot:1986vk}. 
While the desired constraint set $\S$ is the intersection of $\S_1$ and $\S_2$, we approximate the projection on $\cl S$ by $\proj_{\cl S}\simeq \proj_{\cl S_1} \circ \proj_{\cl S_2}$. 
The correct treatment of $\proj_{\cl S}$ would normally require to iteratively combine $\proj_{\cl S_1}$ and $\proj_{\cl S_2}$ (\eg running Generalized Forward-Backward splitting \cite{raguet-gfb} until convergence). 
In our experiments, this approximation did not lead to differences in the estimated patches.

Alg.~\ref{alg:l1} summarizes the different steps involved in the resolution scheme.
It requires a bound $\Gamma$ on the operator norm $\bs K\in\R^{(M+N)\times 2N}$, \ie on $\|\bs K\|=\max_{\bs x:\,\|\bs x\|=1} \|\bs K\bs x\|$.
This is obtained by observing that $\| \Wave \|^2 = 1$ with a proper rescaling due to energy conservation constraints, leaving:
\begin{equation}
\| \bs K \|^2 = \| \big(\begin{smallmatrix}\bs \LD&0\\0&\Wave\end{smallmatrix}\big)\|^2 \leqslant \| \bs \LD \|^2 + 1,
\end{equation}
where $\|\bs \LD\|$ can be efficiently estimated without any spectral decomposition of $\bs \LD$ by using the power method~\cite{Sidky:2011ud}.

While \eqref{eq:pd_functional} may seem unnecessarily complicated at first because it involves both a minimization and a maximization subproblems, the resolution scheme is actually very efficient: it is a first-order method that involves mostly pointwise normalization and thresholding operations.

\begin{algorithm}[H]
	\caption{Primal-dual $\ell_1$ sparse patch reconstruction.}
 	\label{alg:l1}
	\begin{algorithmic}[1]
 	\STATE Take $\Gamma \geqslant \| \bs K \|_2$, choose $\tau, \sigma, \theta$ such that $\Gamma^2\sigma\tau \leqslant 1$, $\theta \in [0, 1]$ and $n$ the number of iterations
	\STATE Initialize: $\bs x^{(0)},\tilde{\bs x}^{(0)} \leftarrow \bs 0$, and $\bs r^{(0)},\bs s^{(0)} \leftarrow \bs 0$
 		\FOR{$i = 0$ to $n-1$}
			\STATE $\bs r^{(i+1)} \leftarrow \prox_{\sigma F_1^*}(\bs r^{(i)} + \sigma \bs \LD \tilde{\bs x}^{(i)})$
			\STATE $\bs s^{(i+1)} \leftarrow \prox_{\sigma F_2^*}(\bs s^{(i)} + \sigma \Wave \tilde{\bs x}^{(i)})$

			\STATE $\bs x^{(i+1)} \leftarrow \proj_\S (\bs x^{(i)} - \frac{\tau}{2} \bs \LD^T \bs r^{(i+1)} - \frac{\tau}{2} \Wave^T \bs s^{(i+1)})$\vspace{.8mm}
			
			\STATE $\tilde{\bs x}^{(i+1)} \leftarrow \bs x^{(i+1)} + \theta\,(\bs x^{(i+1)} - \bs x^{(i)})$
		\ENDFOR
		\RETURN $\hat{\bs p} \leftarrow \bs x^{(n)}$.
	\end{algorithmic}
\end{algorithm}

\subsection{Iterative binary descriptor reconstruction}

To our surprise, when implementing and testing Alg.~\ref{alg:l1} it turned out that it was able to reconstruct not only real-valued descriptors but also binarized ones, \ie it still worked without modifications for some $\bar{\bs p} \in \{-1,1\}^M$ instead of $\R^M$.
This is probably due to the choice of the $\ell_1$-norm in the data term, which tends to attach more importance to the sign of the error than to its actual value.
However, the behavior of our solver in the binarized descriptor case was unstable and it consistently failed to reconstruct some image patches, yielding a null solution.
Hence, we chose to leverage some recent results from 1-bit Compressive Sensing~\cite{Baraniuk:2011wk} to work out a dedicated binary reconstruction scheme.

Keeping the same inverse-problem approach, we substantially modified the functional of \eqref{eq:real_inverse_problem} in two ways:
\begin{enumerate}
\item the data term was changed to enforce bitwise consistency between the LBD computed from the reconstructed patch and the input binary descriptor;
\item to apply the same Binary Iterative Hard Thresholding (BIHT) algorithm as \cite{Baraniuk:2011wk}, we take as sparsity measure the $\ell_0$-norm of the wavelet coefficients instead of the relaxed version obtained with the $\ell_1$-norm.
\end{enumerate}

We are interested by the solution of this new Lasso-type program \cite{tibshirani1996regression}:
\begin{equation}
	\hat{\bs p} = \argmin_{\bs x \in \R^N} \cl J(\bs x)\ \st \| \bs W \bs x \|_0 \leqslant k ~\text{and}~ \bs x \in \S,
	\label{eq:binary_inverse_problem}
\end{equation}
where the constraints enforces both the validity and the $k$-sparsity of $\bs x$ in the wavelet domain.

Inspired by \cite{Baraniuk:2011wk}, we set our data term as
\begin{equation}
\cl J(\bs x) = \| [\,\bar{\bs p} \odot \B(\LD \bs x) ]_-\|_1.
\label{eq:J-bin}
\end{equation}
Qualitatively, $\cl J$ measures the LBD consistency of $\bs x$ with $\bs p$, with $\cl J(\bs x)=0$ iff $\cl B(\bs{\cl L}\bs x)=\bar{\bs p}$. Each component of the Hadamard product in the definition of $\cl J$ is either positive (both signs are the same) or negative. Since the negative function sets to~0 the consistent components, the $\ell_1$-norm finally adds the contribution of each inconsistent entry.
Note that at the time of writing we do not know a solution for the proximal mapping associated to this data term $\cl J$, which explains our choice for BIHT over the primal-dual solver used in the previous section.

Similarly to the way Iterative Hard Thresholding aims at solving an $\ell_0$-Lasso problem \cite{blumensath2009iterative},
BIHT finds one solution of \eqref{eq:binary_inverse_problem} by repeating the three following steps until convergence:
\begin{enumerate}
\item computing a step of gradient descent of the data term;
\item enforcing sparsity by projecting the intermediate estimate to the set of patches with at most $K$ non-zero coefficients;
\item enforcing the mean-value constraint on the result.
\end{enumerate}
This last operation was already studied in the previous section for the real-valued case: it is the projection onto the set $\S$.
The $\ell_0$-norm constraint is applied by Hard Thresholding and amounts to keeping the $K$ biggest coefficients of the wavelet transform of the estimate and discarding the others.
We write this operation $\cl H_K$.
Finally, unlike in the primal-dual algorithm, the gradient descent of the data term has to be computed.
The result of Lemma 5 in \cite{Baraniuk:2011wk} applies in our case and a subgradient of the data term in \eqref{eq:binary_inverse_problem} is:
\begin{equation}
  \partial \cl J(\bs x)\ \ni\ \tinv{2} \bs \LD^T \big(\B(\bs \LD \bs x) - \bar{\bs p}\big),
\end{equation}
\ie the back-projection of the binary error.

Putting everything together, we obtain Alg.~\ref{alg:biht} that is the adaptation of BIHT to the reconstruction of image patches from their LBD representation. 
Again, this algorithm is made of simple elementary steps. 
The parameter $\tau=1/M$ guarantees that the current solution $\bs x^{(i)}$ and the gradient step $\frac{\tau}{2} \bs \LD^T \big(\bar{\bs p} - \B(\bs \LD \bs x^{(i)})\big)$ have comparable amplitudes \cite{Baraniuk:2011wk}. 
Since $M$ is determined from the LBD size, only the patch sparsity level $K$ in the wavelet basis must be tuned (see Sec.~\ref{sec:impl-deta}). 
In our experiments, however, the algorithm was not very sensitive to the value of $K$. 

\begin{algorithm}[H]
	\caption{BIHT patch reconstruction.}
 	\label{alg:biht}
 	\begin{algorithmic}[1]
	\STATE Take $\tau = 1/M$, choose $K$ the number of non-zero coefficients and $n$ the number of iterations
	\STATE Initialize: $\bs x^{(0)} \leftarrow 0$ and $\bs a_0 \leftarrow 0$.
 		\FOR{$i = 0$ to $n-1$}
			\STATE $\bs a^{(i+1)} \leftarrow \bs x^{(i)} + \frac{\tau}{2} \bs \LD^T (\bar{\bs p} - \sign(\bs \LD \bs x^{(i)}))$
			\STATE $\bs b^{(i+1)} \leftarrow \cl H_K (\bs W \bs a^{(i+1)})$
			\STATE $\bs x^{(i+1)} \leftarrow \proj_\S \big( \bs W^T \bs b^{(i+1)} \big)$
		\ENDFOR
		\RETURN $\hat{\bs p} \leftarrow \bs x^{(n)}$.
	\end{algorithmic}
 \end{algorithm}
 
\section{Results and discussion}
\label{sec:results}

\subsection{Implementation details}
\label{sec:impl-deta}

For the reconstruction tests presented in this Section, we re-implemented two of the different LBDs: BRIEF and FREAK.
For BRIEF, we chose a uniform distribution for the location of the Gaussian measurements, whose support was fixed to $3 \times 3$ pixels, following the original paper~\cite{Calonder:2010tx}.
For FREAK, we did not take into account the orientation of the image patches (see~\cite{Anonymous:_4iTrXEX}, Sec. 4.4) but we also implemented two variants:
\begin{itemize}
\item EX-FREAK, for EXhaustive-Freak, computes all the possible pairs from the retinal pattern;
\item RA-FREAK, for RAndomized-FREAK, randomly selects its pairs from the retinal pattern.
\end{itemize}

All the operators were implemented in C++ with the OpenCV library\footnote{Freely available at \url{http://opencv.org}} and used the same codebase for fair comparisons, varying only in the measurement pair selection.
The code used to generate the examples in this paper is available online and can be retrieved from the page \url{http://lts2www.epfl.ch/software/}.

The sensing operator was implemented in the following way:
\begin{itemize}
\item the forward operator $\bs \LD$ is obtained through the use of integral images for a faster computation of the Gaussian weighted integrals $\langle \G_{q_i,\sigma_i},\G_{q'_i,\sigma'_i}\rangle$.
This approximation has become standard in feature descriptor implementations since it allows a huge acceleration of the computations, see for example~\cite{Bay:2006ug};
\item the backward operator $\bs \LD^T$ is the combination of the frame vectors of the considered LBD weighted by the input vector of coefficients.
Hence, we avoid explicitly forming $\bs \LD^T$ by computing on the fly the image representation of each vector $\bs \LD_i$ of \eqref{eq:Li}.
\end{itemize}


In the previous sections, we have proposed two algorithms that aimed at reconstructing an image patch from the corresponding local descriptor.
In order to assess their relevance and the quality of the reconstructions we have applied them on whole images according to the following protocol:
\begin{enumerate}
\item an image is divided into patches of size $\sqrt{N} \times \sqrt{N}$ pixels, with an horizontal and vertical offset of $N_{\rm off}$ pixels between each patch;
\item each patch is reconstructed independently from its LBD representation using the additional constraint that its mean should be $0.5$, \ie the mean of the input dynamic range;
\item reconstructed patches are back-projected to their original image position.
Wherever patches overlap, the final result is simply the average of the reconstructions.
\end{enumerate}
Hence, the experiments introduce an additional parameter which is the offset between the selected patches.

Note that in contrast with~\cite{Weinzaepfel:2011jh} our methods do not require the use of a seamless patch blending algorithm.
Simple averaging does not introduce artifacts.
Also, we do not require the knowledge of the scale and orientation of the keypoint to reconstruct: we assume it is a patch of fixed size aligned with the image axes.
This is in line with the 1-bit feature detection and extraction pipeline: the genuine FAST detector does not consider scale and orientation, and the BRIEF descriptor is not rotation or scale-invariant.
Later descriptors such as FREAK were trained in an affine-invariant context; hence by considering only fixed width and orientation our algorithm is suboptimal.

We used patches of $32 \times 32$ pixels, LBDs of 512 measurements and run Alg.~\ref{alg:l1} and Alg.~\ref{alg:biht} for 1000 and 200 iterations respectively.
In Alg.~\ref{alg:l1}, the trade-off parameter $\lambda$ was set to $0.1$.
We tried different values for the sparsity $K$ in Alg.~\ref{alg:biht} (retaining between 10\% and 40\% of the wavelet coefficients) but the results did not vary in a meaningful way.
Thus we fixed $K$ throughout all the experiments to keep the 40\% greater coefficients of $\Wave \p$, choosing the Haar wavelet as analysis operator.

\begin{figure}[htbp]
	\begin{center}
		\subfigure[Lena]{\includegraphics[width=.25\linewidth]{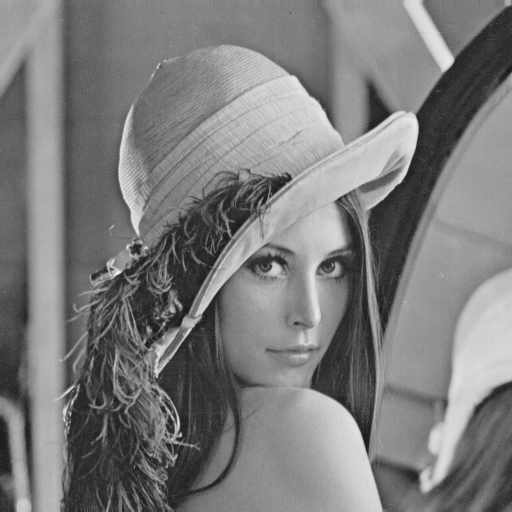}} \quad
		\subfigure[Barbara]{\includegraphics[width=.25\linewidth]{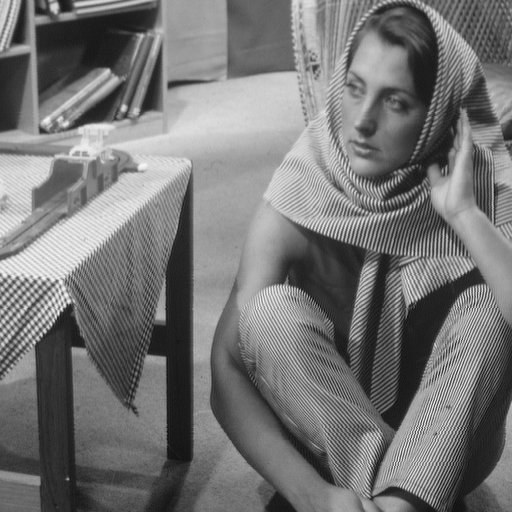}} \quad
		\subfigure[Kata]{\includegraphics[height=.25\linewidth]{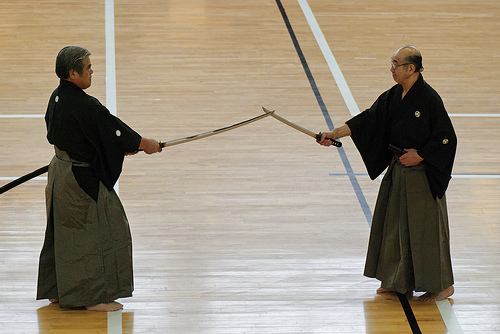}}
	\caption{Original images and designated name in the text.}
	\label{fig:originals}
	\end{center}
\end{figure}

The original \verb+Lena+, \verb+Barbara+ and \verb+Kata+ images can be seen in Fig.~\ref{fig:originals}.

\subsection{Reconstruction results}
\label{sec:reconstr-results}

\begin{figure*}[htbp]
	\begin{center}
		\subfigure[BRIEF]{\includegraphics[width=.35\textwidth]{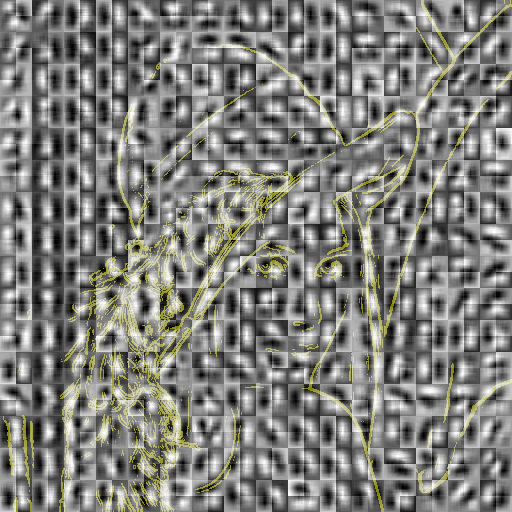}} \quad
		\subfigure[FREAK]{\includegraphics[width=.35\textwidth]{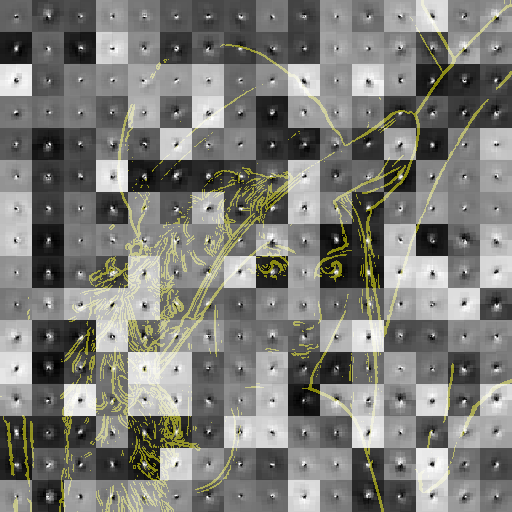}} \quad
		\subfigure[RA-FREAK]{\includegraphics[width=.35\textwidth]{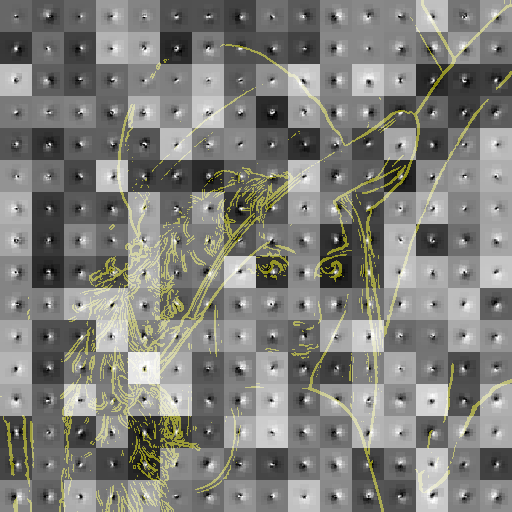}} \quad
		\subfigure[EX-FREAK]{\includegraphics[width=.35\textwidth]{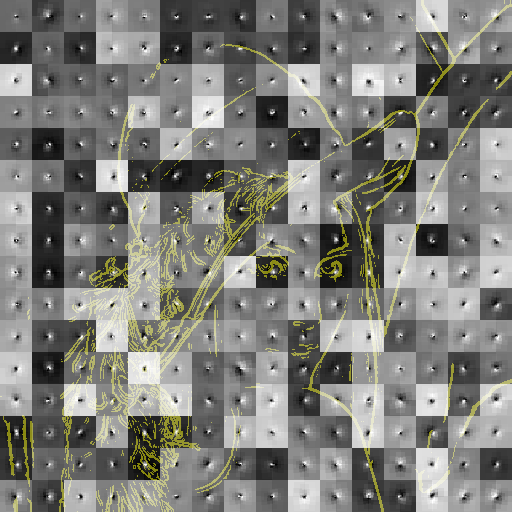}}
	\caption{Reconstruction of Lena from binary LBDs using Alg. \ref{alg:biht}. 
			There is no overlap between the patches used in the experiment, thus giving a blockwise aspect.
			We have overlaid some edges from the original image.
			In each case, the orientation selected for the output patch is consistent with the original main gradient direction.
			Note also the difference between BRIEF (random measurements spread over an image patch) and FREAK and its derivatives (fine measurements with higher density near the center of the patch): 
			the former gives large blurred edges covering the whole patch, while the latter affect the dominant gradient direction to the central pixel, leaving the periphery almost untouched.}
	\label{fig:lena-zero-overlap}
	\end{center}
\end{figure*}

At first glance, the reconstruction results for non-overlapping patches visible in Fig.~\ref{fig:lena-zero-overlap} seem very weird and have sometimes very little in common with the original image.
However, if we overlay the original edges on top of the reconstructed images, one can see that each estimated patch contains a correct version of the original gradient direction.
This shows that all the four LBDs that we have experimented capture the local gradient and that this information is enough for Alg.~\ref{alg:biht} to infer the original value.
Even curved lines and cluttered area are encoded by the binary descriptors: see the shoulder of Lena and the feathers of her hat (Fig.~\ref{fig:lena-zero-overlap}).
While there is a significant difference between the reconstruction from BRIEF and FREAK, the variants RA-FREAK and EX-FREAK lead to results almost identical with the original FREAK.

\begin{figure}[htbp]
	\begin{center}
		\subfigure[$N_{\rm off}=32$ pixels]{\includegraphics[width=.23\textwidth]{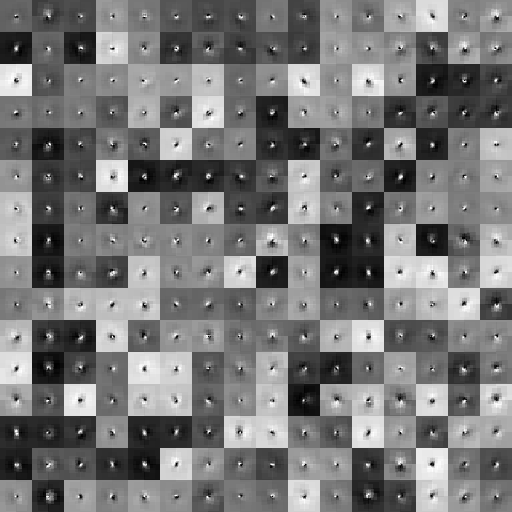}} \quad
		\subfigure[$N_{\rm off}=16$ pixels]{\includegraphics[width=.23\textwidth]{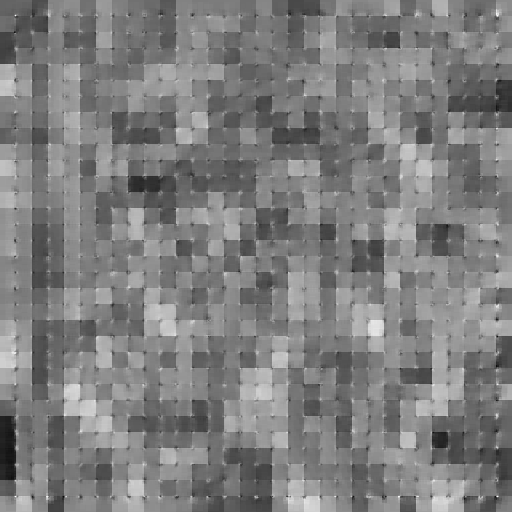}} \quad
		\subfigure[$N_{\rm off}=8$ pixels]{\includegraphics[width=.23\textwidth]{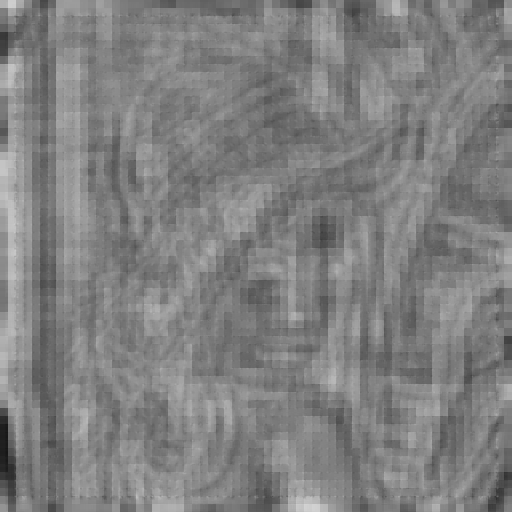}} \quad
		\subfigure[$N_{\rm off}=1$ pixel]{\includegraphics[width=.23\textwidth]{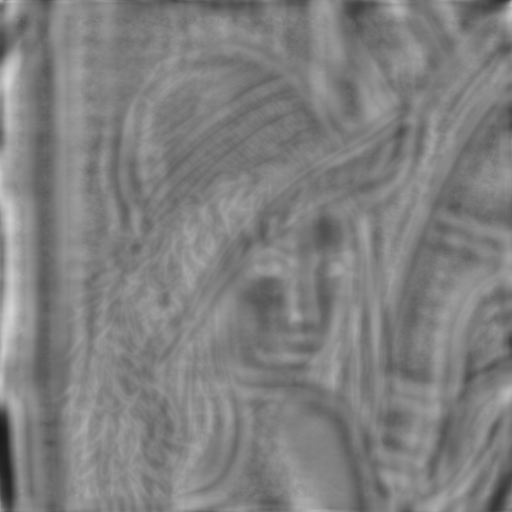}}
	\caption{Reconstruction of Lena from binary FREAKs.
			The size of the patches is kept fixed at $32 \times 32$ pixels, while their spacing is gradually reduced.
			We start with an offset of 32 pixels, \ie no overlap, until a dense reconstruction.
			In the limit when each pixel is reconstructed from its neighborhood the individual edge bits chain up and one can clearly distinguish the original image contours,  like after the application of a Laplacian filter.}
	\label{fig:lena-overlaps}
	\end{center}
\end{figure}

Keeping the patch size constant at $32 \times 32$ pixels, some results for various offsets between the patches can be seen in Fig.~\ref{fig:lena-overlaps}.
An increased number of overlapping patches does dramatically improve the quality of the reconstruction using FREAK.
This can be understood easily by considering the peculiar shape of the reconstruction without overlap: each estimated patch contains the correct gradient information at its center only.
By introducing more overlap between the patches these small parts of contour sum up to recreate the original objects.

\begin{figure*}[htbp]
	\begin{center}
		\subfigure{\includegraphics[width=.25\linewidth]{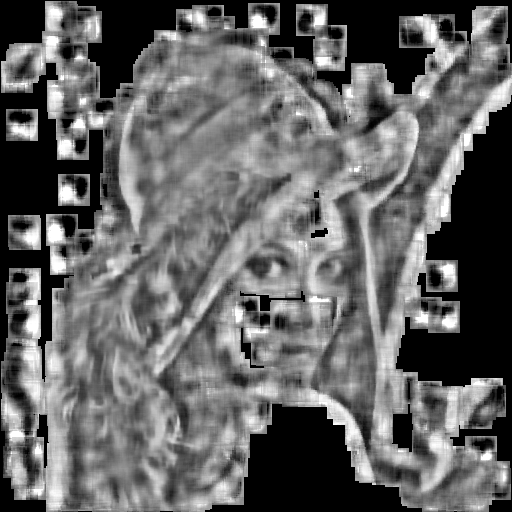}} \quad
		\subfigure{\includegraphics[width=.25\linewidth]{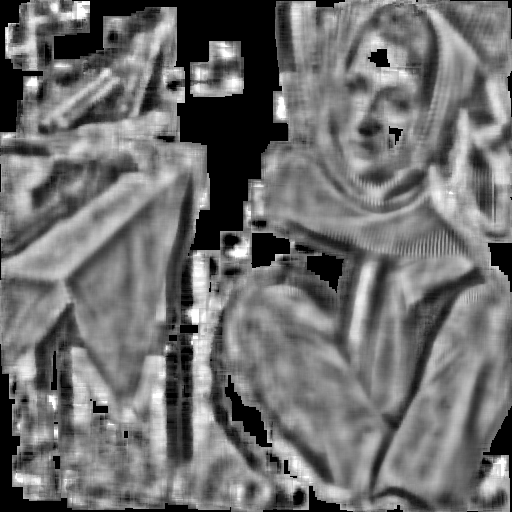}} \quad
		\subfigure{\includegraphics[height=.25\linewidth]{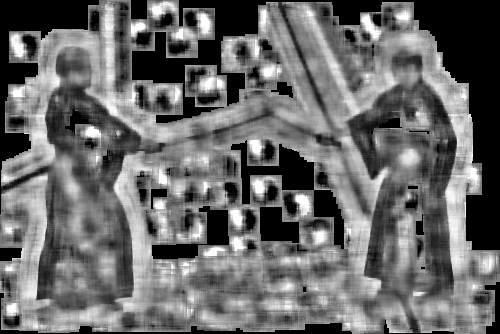}}
	\caption{Reconstruction of floating-point (non binarized) BRIEFs centered on FAST keypoints.}
	\label{fig:real-desc}
	\end{center}
\end{figure*}

\begin{figure*}[htbp]
	\begin{center}
		\subfigure{\includegraphics[width=.25\linewidth]{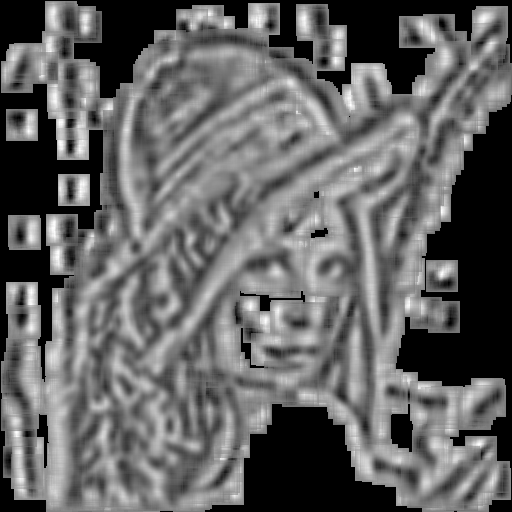}} \quad
		\subfigure{\includegraphics[width=.25\linewidth]{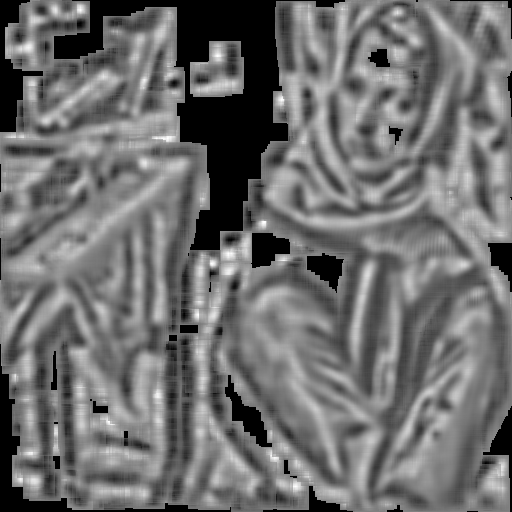}} \quad
		\subfigure{\includegraphics[height=.25\linewidth]{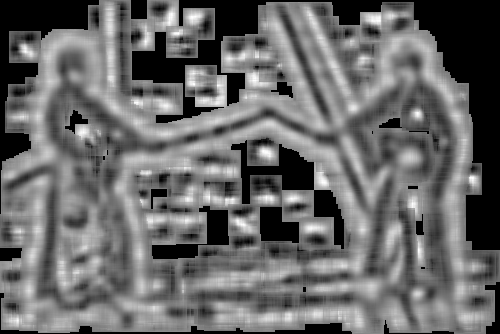}} \\

		\subfigure{\includegraphics[width=.25\linewidth]{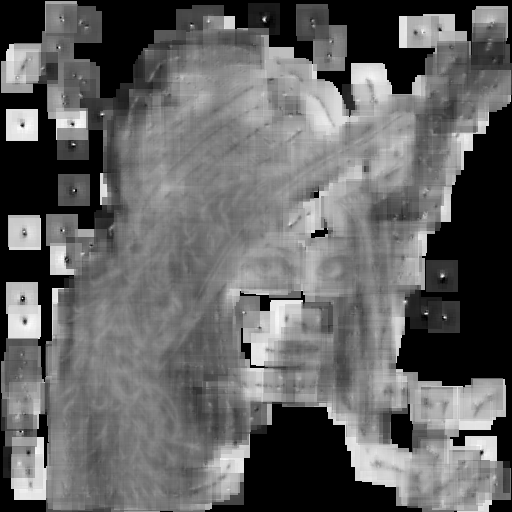}} \quad
		\subfigure{\includegraphics[width=.25\linewidth]{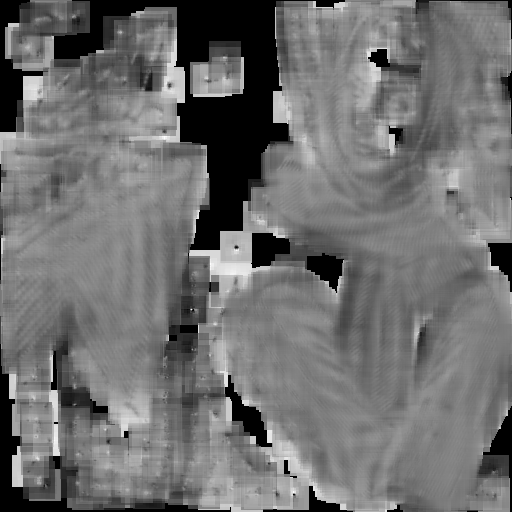}} \quad
		\subfigure{\includegraphics[height=.25\linewidth]{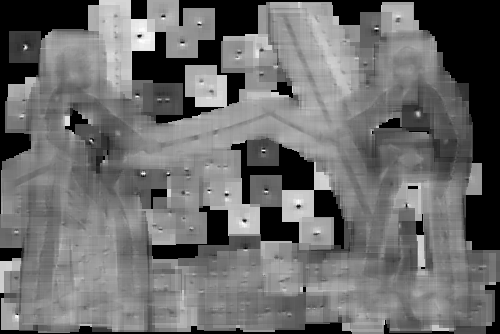}}
	\caption{Reconstruction of LBDs centered on FAST keypoints only.
			\emph{Top row:} using BRIEF.
			\emph{Bottom row:} using FREAK.
			Since the detected points are usually very clustered there is a dense overlap between patches, yielding a visually plausible reconstruction.
			The original image content has been correctly recovered by Alg.~\ref{alg:biht} from binary descriptors, and eavesdropping the communications of a mobile camera (\eg embedded in a smartphone) could reveal private data.}
	\label{fig:fast}
	\end{center}
\end{figure*}

\begin{figure*}[htbp]
	\begin{center}
		\subfigure[Original image]{\includegraphics[height=.36\linewidth]{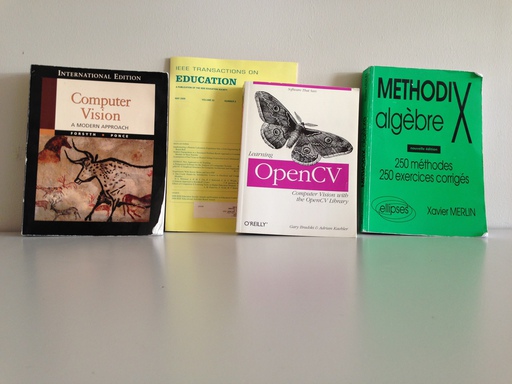}} \quad
		\subfigure[FAST + Floating-point BRIEF]{\includegraphics[height=.36\linewidth]{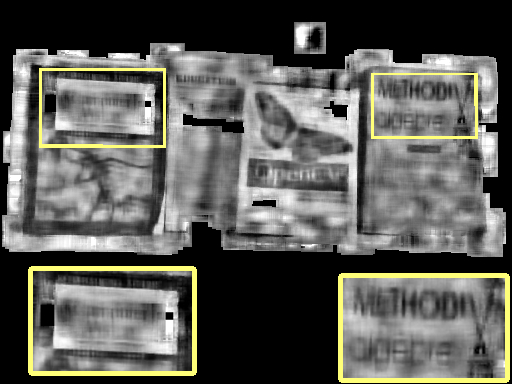}} \quad
		\subfigure[FAST + binarized BRIEF]{\includegraphics[height=.36\linewidth]{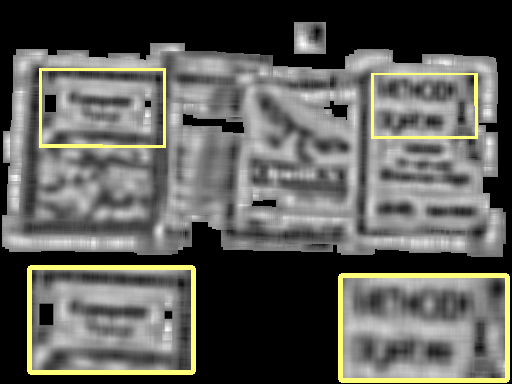}} \quad
		\subfigure[FAST + binarized FREAK]{\includegraphics[height=.36\linewidth]{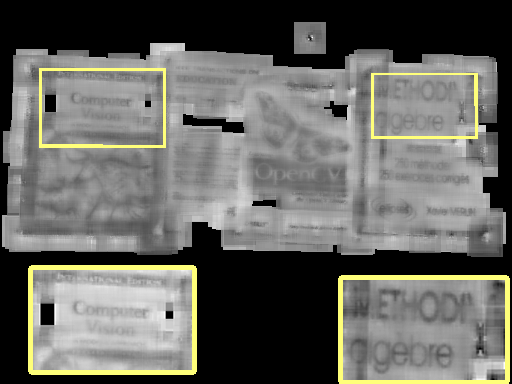}}
	\caption{Reconstruction of book covers.
	The bottom part of each image shows in inset a close-up view of two book titles.
	This experiment confirms the difference between BRIEF and FREAK: while the former extracts salient shapes such as the auroch and the butterfly, the latter is more successful at reconstructing the text.
	Note that FREAK allows to read 3 titles out of 4, hence demonstrating the potential existing privacy breach in case of mobile communications eavesdropping.}
	\label{fig:books}
	\end{center}
\end{figure*}

Instead of computing patches at fixed positions and offsets, an experiment more relevant with respect to privacy concerns consists in reconstructing only the patches associated with an interest point detector.
For the results shown in Fig.~\ref{fig:real-desc}, we have first applied the FAST feature detector of OpenCV with its default parameters and discarded the remaining part of the image, hence the black areas, and used real-valued descriptors.
Fig.~\ref{fig:real-desc} shows the results of the same experiment with binary descriptors.
Since FAST keypoints tend to aggregate near angular points and corners, this lead to a relatively dense reconstruction.
In each of the three images, the original content can clearly be recognized and a great part of the background clutter has been removed.
Thus, one can add as a side note that FAST keypoints are a good indicator of image content saliency.
The results in Fig.~\ref{fig:fast} and Fig.~\ref{fig:books} extend to binary descriptors an important privacy issue that was raised before by \cite{Weinzaepfel:2011jh} for SIFT: if one can intercept keypoint data sent over a network (\eg to an image search engine), then it is possible to find out what the legitimate user was seeing.

\begin{figure*}[htbp]
	\begin{center}
		\subfigure{\includegraphics[width=.17\linewidth]{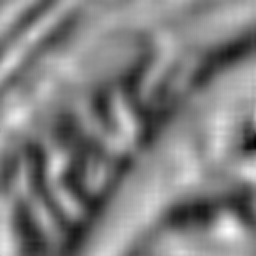}} \quad
		\subfigure{\includegraphics[width=.17\linewidth]{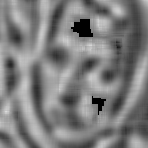}} \quad
		\subfigure{\includegraphics[width=.17\linewidth]{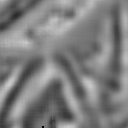}} \quad
		\subfigure{\includegraphics[width=.17\linewidth]{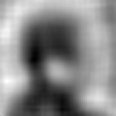}} \quad
		\subfigure{\includegraphics[width=.17\linewidth]{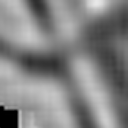}} \\
		\subfigure{\includegraphics[width=.17\linewidth]{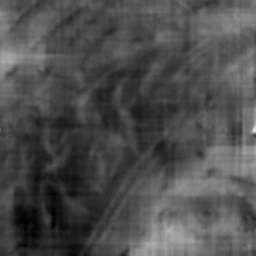}} \quad
		\subfigure{\includegraphics[width=.17\linewidth]{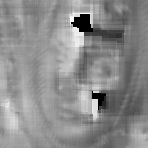}} \quad
		\subfigure{\includegraphics[width=.17\linewidth]{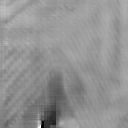}} \quad
		\subfigure{\includegraphics[width=.17\linewidth]{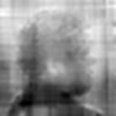}} \quad
		\subfigure{\includegraphics[width=.17\linewidth]{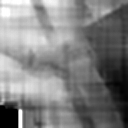}} \\
	\end{center}
	\caption{Details of the reconstructions from Fig.~\ref{fig:fast}.
			\emph{Top row:} using BRIEF as LBD. \emph{Bottom row:} using FREAK.
			The reconstructed patches were selected by the application of the FAST detector with identical parameters.
			While BRIEF is successful at capturing large gradient orientations, hence giving pleasant results when the image is seen from a distance, FREAK captures more accurate orientations in the center of the patches.
			Thus finer details are recovered: notice for example the eyes in the pictures of Lena and Barbara, the textures from Barbara or the face and the fingers in the kata image.
			For this Figure, some additional contrast enhancement post-processing was applied to emphasize the point.}
	\label{fig:brief-vs-freak-detail}
\end{figure*}

\begin{figure}[htbp]
	\begin{center}
		\subfigure[BRIEF, weighted values]{\includegraphics[width=.21\textwidth]{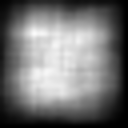}} \quad
		\subfigure[FREAK, weighted values]{\includegraphics[width=.21\textwidth]{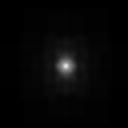}} \\
		\subfigure[BRIEF, occurrence count]{\includegraphics[width=.21\textwidth]{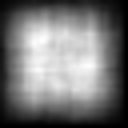}} \quad
		\subfigure[FREAK, occurrence count]{\includegraphics[width=.21\textwidth]{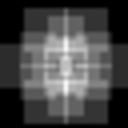}} \\
	\end{center}
	\caption{Comparison of the spatial weights in BRIEF and FREAK basis functions.
			In the top row, we display the sum of the absolute values of the weight of each pixel when computing a descriptor.
			Brighter means higher importance.
			One can see that BRIEF considers almost equally pixels all over the patch, while FREAK gives a very high weight to the centre.
			The bottom row shows how many times a pixel value was read to generate the description vector.
			Here, brighter means often retained.
			This shows that FREAK uses peripheral values, but with a low ponderation.
			BRIEF is clearly more democratic since the weight pattern is similar to the occurrence pattern.}
	\label{fig:brief-vs-freak-bases}
\end{figure}

Reconstruction results from BRIEF and FREAK are strikingly different (Fig.~\ref{fig:lena-zero-overlap}).
While BRIEF leads to large, blurred edge estimates that occupy almost entirely the original patch, FREAK produces small accurate edges almost confined in the center of the patch.
This allows us to point at a fundamental difference between BRIEF and FREAK.
While the former does randomly sample a rough estimate of the dominant gradient in the neighborhood, the latter concentrates its finest measurements and allows more bits (Fig.~\ref{fig:brief-vs-freak-bases}) to the innermost part.
Thus, the inversion of BRIEF leads to a fuzzy blurred edge dividing two areas since the information is spread spatially over the whole patch, while the reconstruction of FREAK produces a small but accurate edge surrounded by a large low-resolution area.
This is confirmed by the experiments shown in Fig.~\ref{fig:brief-vs-freak-detail}.
One can especially remark the eyes of \verb+Lena+ and \verb+Barbara+ and the crossed pattern of the table blanket from \verb+Barbara+ which exhibit fine details using FREAK that are missing with BRIEF.
In the \verb+Kata+ image, one can almost recognize the face of the characters with FREAK, while the fingers holding the sword are clearly distinguishable.

Figure \ref{fig:brief-vs-freak-bases} compares the measurement strategies of BRIEF and FREAK for 512 measurements.
The top row displays the sum of the absolute values of the weights applied to a pixel when computing the descriptor, \ie $\sum_{i=1}^M |(\G_{\bs q_i,\sigma_i})_j| + |(\G_{\bs q_i',\sigma_i'})_j|$ in \eqref{eq:L} for the $N$ pixels $1\leq j\leq N$.
We clearly see that BRIEF measures patch intensity almost uniformly over its domain, while FREAK focuses its patch observation on the patch domain center. Yet, when plotting in how many LBD measurements a pixel contributed (Fig.~\ref{fig:brief-vs-freak-bases}, bottom row) one can see that FREAK also uses peripheral pixels. Since both the weight and occurrence patterns are similar with BRIEF, it means that this LBD is democratic and gives all the pixels a similar importance.


\begin{figure}[htbp]
	\begin{center}
		\subfigure[BRIEF, $M=128$]{\includegraphics[width=.20\textwidth]{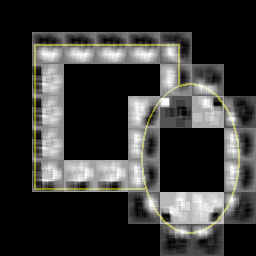}} \quad
		\subfigure[FREAK, $M=128$]{\includegraphics[width=.20\textwidth]{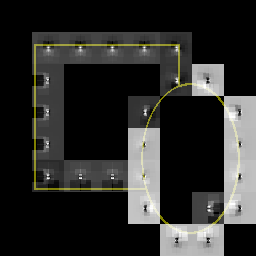}}
	\caption{Zero-overlap reconstruction of a synthetic image using LBDs of 128 measurements instead of 512 a sin the other experiments (and 256 in most image matching softwares).
			Note that in spite of the huge information loss (compression ratio of 256:1 for each patch) the directions of the edges are correctly estimated.}
	\label{fig:measures}
	\end{center}
\end{figure}

\begin{figure*}[htbp]
	\begin{center}
		\subfigure{\includegraphics[width=.325\linewidth]{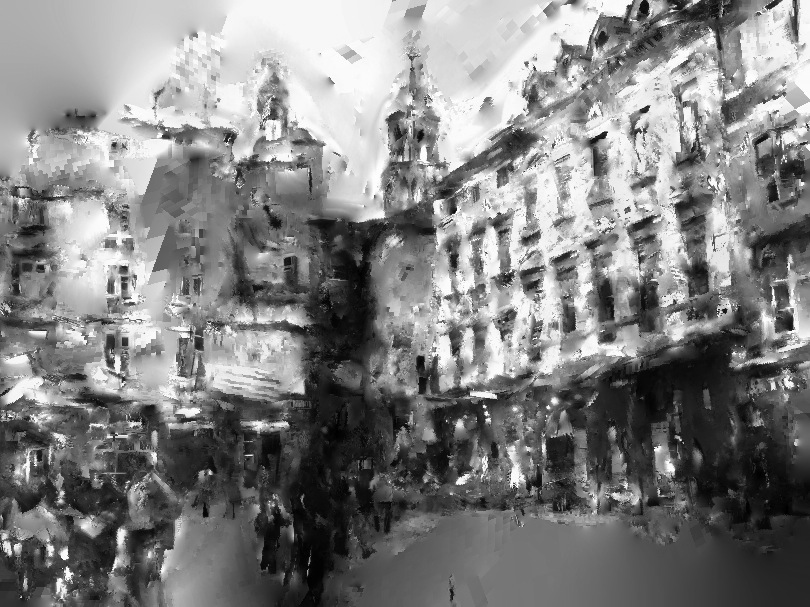}} \hfill
		\subfigure{\includegraphics[width=.325\linewidth]{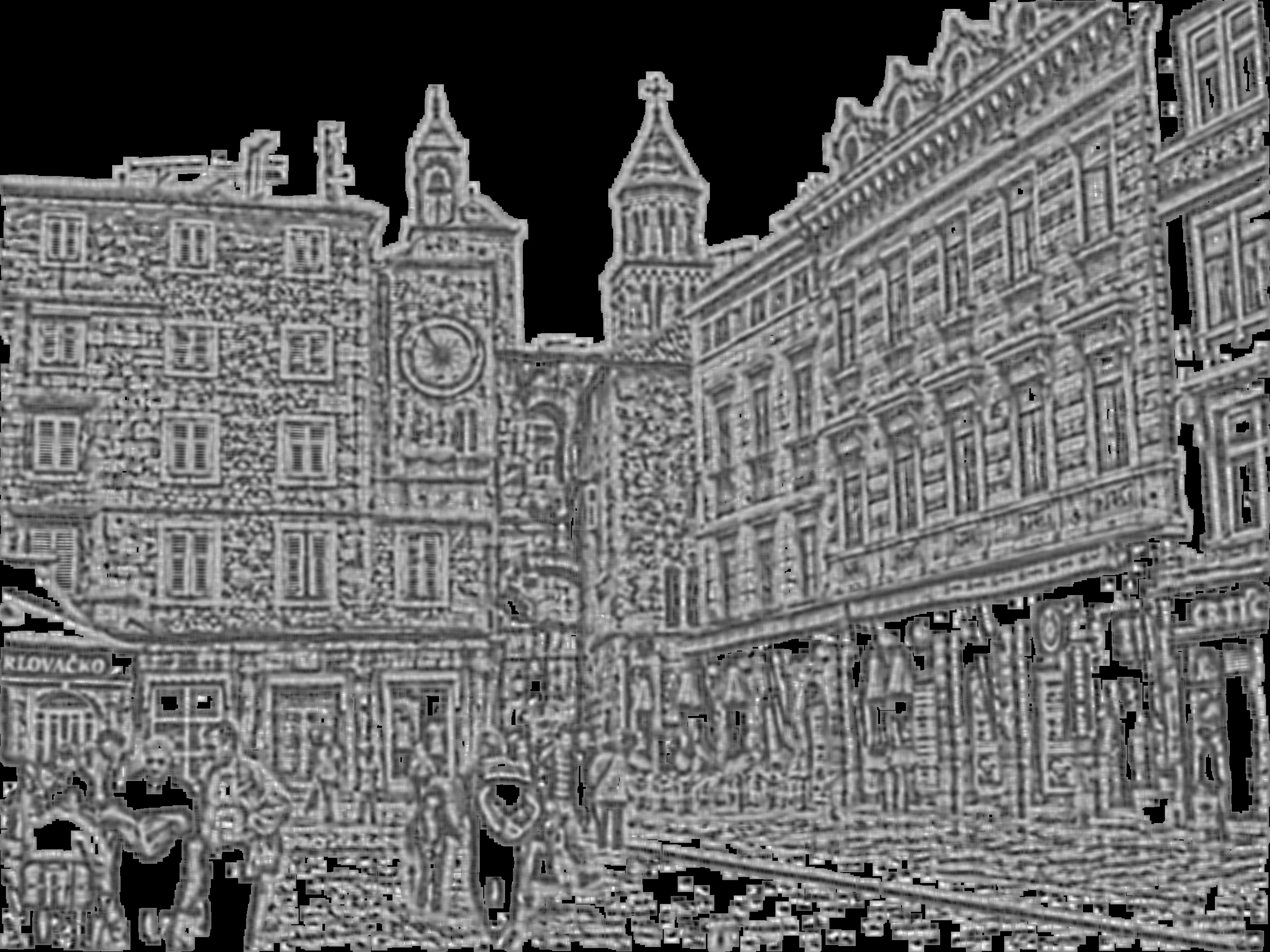}} \hfill
		\subfigure{\includegraphics[width=.325\linewidth]{215100-brief-fast-p32}}
		\subfigure{\includegraphics[width=.325\linewidth]{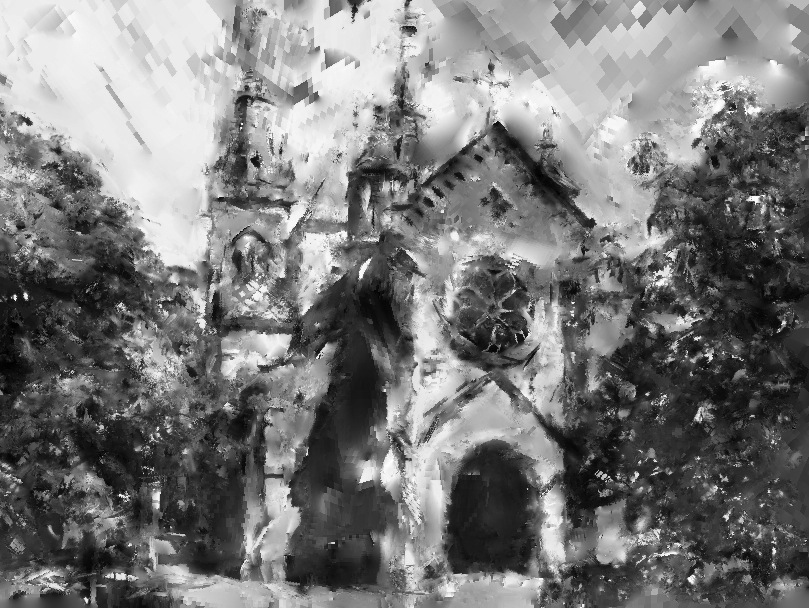}} \hfill
		\subfigure{\includegraphics[width=.325\linewidth]{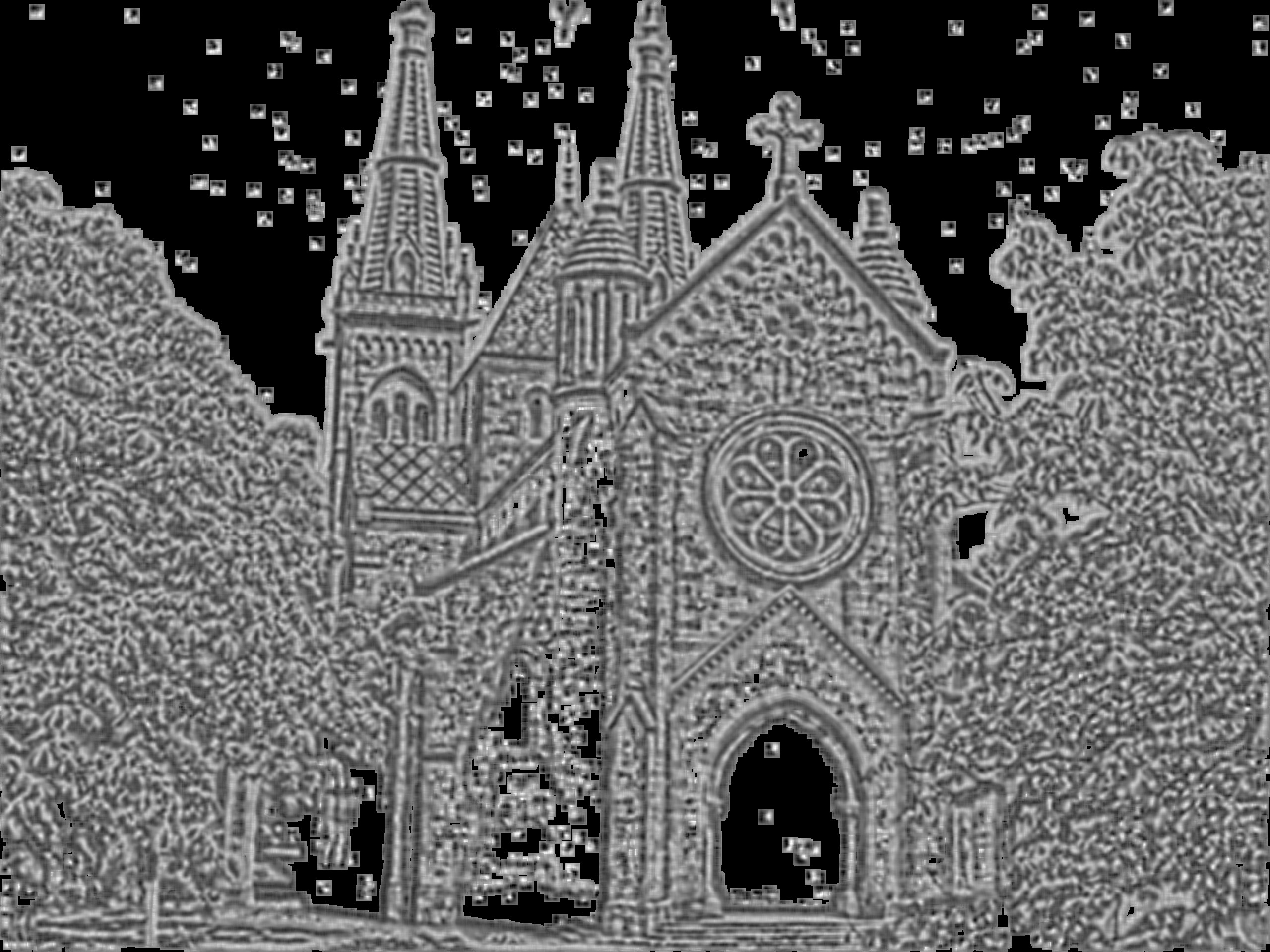}} \hfill
		\subfigure{\includegraphics[width=.325\linewidth]{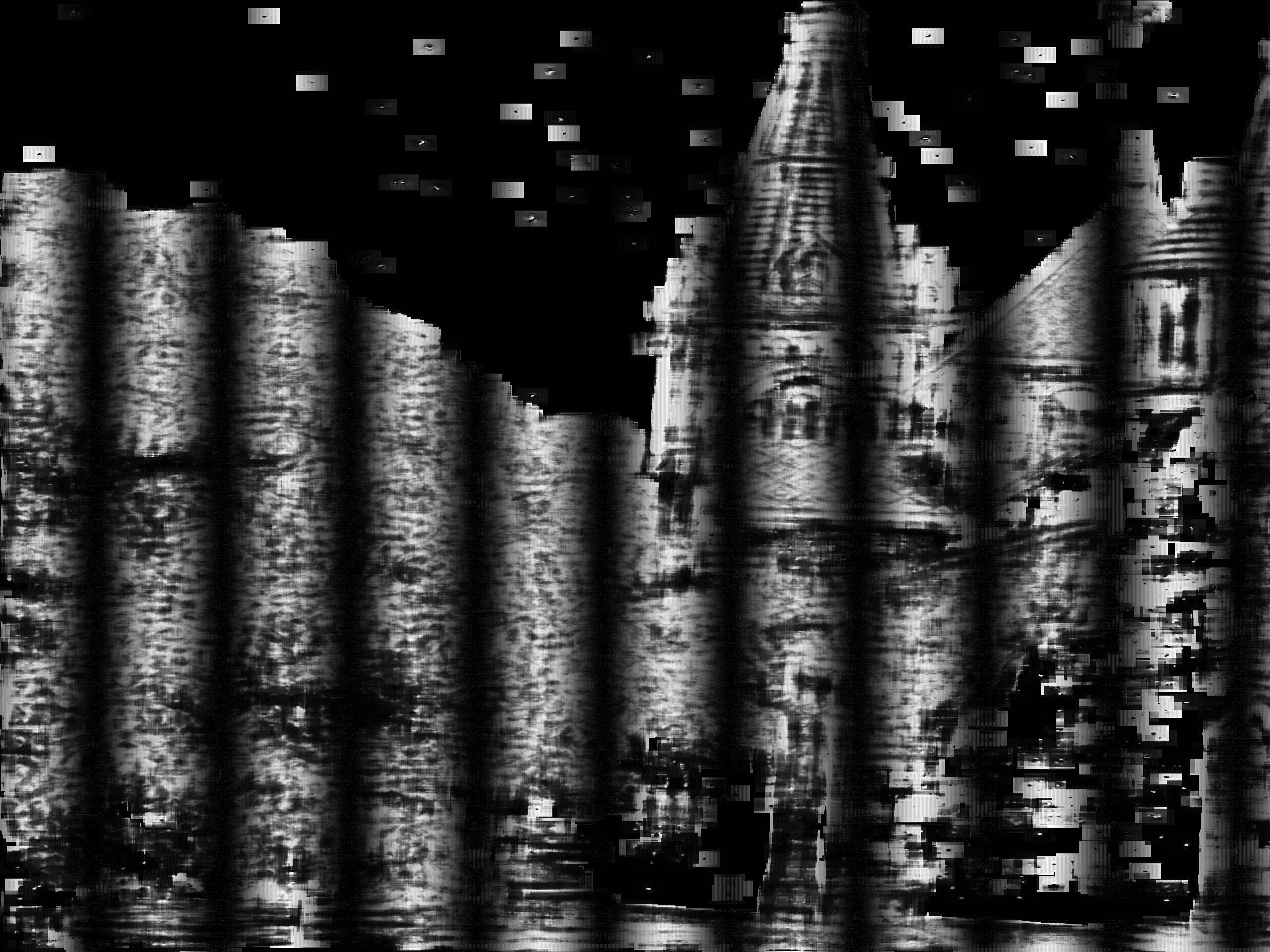}}
		\subfigure{\includegraphics[width=.325\linewidth]{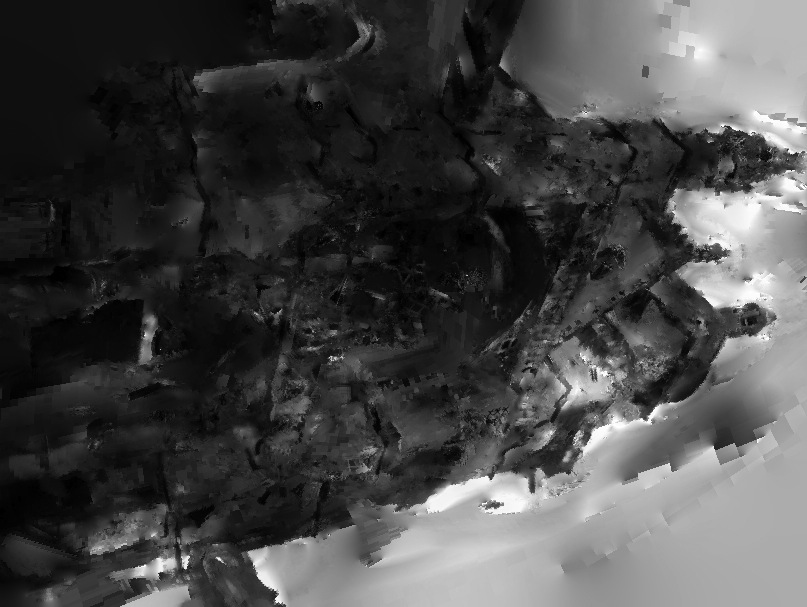}} \hfill
		\subfigure{\includegraphics[width=.325\linewidth]{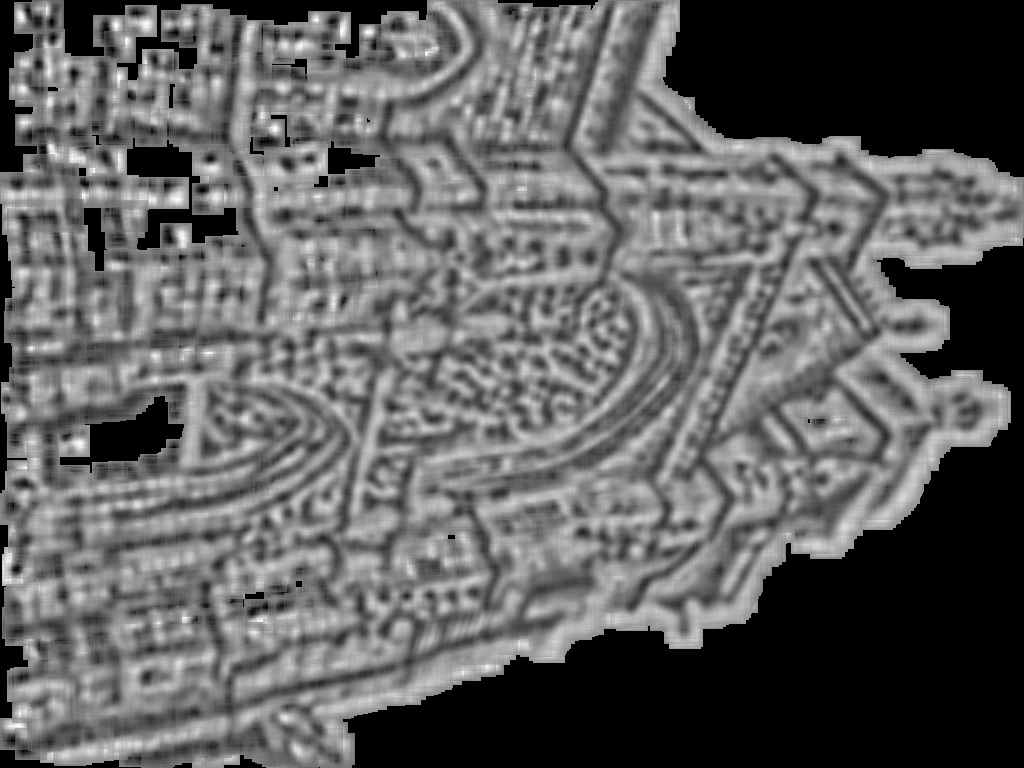}} \hfill
		\subfigure{\includegraphics[width=.325\linewidth]{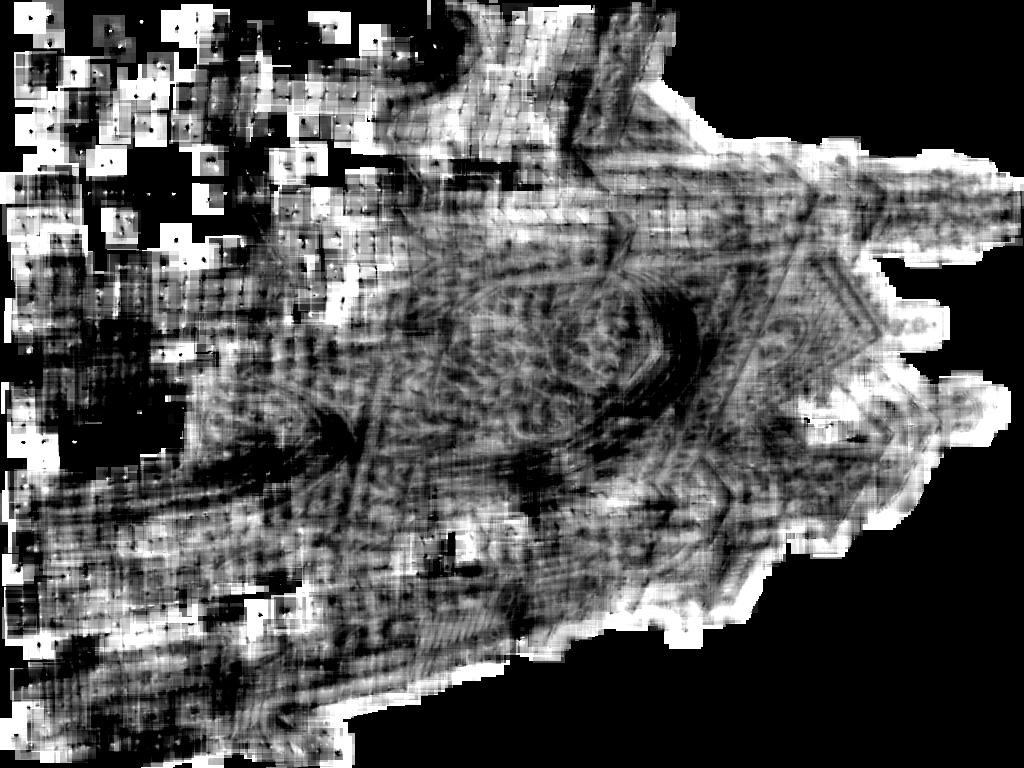}}
	\caption{Comparison with~\cite{Weinzaepfel:2011jh}.
	From left to right: image reconstructed with the method of~\cite{Weinzaepfel:2011jh}, our binary reconstruction algorithm using FAST+BRIEF (middle) and FAST+FREAK (right).
	We used patches of $32 \times 32$ pixels in our algorithm.
	The contrast of the FREAK results was enhanced for readability, but this Figure is best viewed online in electronic version.}
	\label{fig:comparison}
	\end{center}
\end{figure*}

\subsection{Quality and stability of the reconstruction}
\label{sec:qual-stab-reconstr}

Because of the very peculiar structure of the LBD operators, establishing strong mathematical properties on these matrices is a very arduous task, especially in the 1-bit case.
As a consequence, finding indubitable theoretical grounds to the success of our BIHT reconstruction algorithm still remains to be investigated.
Intuitively, one can however remark that the conditions used to ensure the existence of a reconstruction in Compressive Sensing, like the famous Restrictive Isometry Property (RIP), are only \emph{sufficient} conditions and are by no means necessary conditions.
Since LBDs were designed to accurately describe some image content, they are probably more efficient than random sensing matrices.
Hence, they can capture more information from an input patch with very few measurements and with a more brutal quantization at the cost of a loss in genericity: they are specialized sensors tuned to image keypoints.

An important parameter with respect to the expected quality of the reconstructions is of course the length $M$ of the LBDs.
Since we lack a reliable quality metric to assess the reconstructions, we have proceeded to visual comparisons between the original image contours and the reconstructed gradient directions on a synthetic image.
As can be seen in Fig.~\ref{fig:measures}, dominant orientations are reconstructed correctly until $M = 128$ measures.
Smaller sizes yield blocky estimated patches where it is hard to infer any edge direction.



\section{Conclusion and future work}

In this work, we have presented two algorithms that can successfully reconstruct small image parts from a subset of local differences without requiring external data or prior training.
Both algorithms leverage an inverse problem approach to tackle this task and use as regularization constraint the sparsity of the reconstructed image patches in some wavelet frame.
They rely however on different frameworks to solve the corresponding problem.

The first method relies on proximal calculus to minimize a convex non-smooth objective function, adopting a deconvolution-like approach.
While this functional was not specifically designed for 1-bit LBDs, it  has proved to be robust enough to provide some 1-bit reconstructions, but it does lack stability in this case.
On the other hand, the second method was built from the ground up to handle 1-bit LBDs, and thus provides stable results.
The reconstruction process is guided by a hard sparsity constraint in the wavelet domain.

There are several levels to exploit and interpret our results.
First, they can have an important industrial impact.
Since it is possible to easily invert LBDs without additional information, mobile application developers cannot simply move from SIFT to LBDs in order to avoid the privacy issues raised by~\cite{Weinzaepfel:2011jh}.
Hence, they need to add an additional encryption tier to their feature point transmission process if the conveyed data can be sensitive or private.

Second, the differences in the reconstruction from different LBDs can help researchers to design their own LBDs.
For example, our experiments have pointed out that BRIEF encodes information at a coarser scale than SIFT, and maybe both descriptors could be combined in some way to create a scale-aware descriptor taking advantage of both patterns.
Hence, our work can be used as a tool to study and compare binary descriptors providing a different information than standard matching benchmarks.
Furthermore, the fact that real-value differences yield comparable results as binarized descriptors legitimates \emph{a posteriori} the performance of LBDs in matching benchmarks: they encode most of the originally available information.
 
Finally, our framework for 1-bit contour reconstruction could be combined with the previously proposed Gradient Camera concept~\cite{Tumblin:2005wm}, leading to the development of a 1-bit Compressive Sensing Gradient Camera.
This disruptive device would ally the qualities of both worlds with an extended dynamic range and low power consumption.
Exploiting the retinal pattern of FREAK and our reconstruction framework could also yield neuromorphic cameras mimicking the human visual system that could be useful for medical and physiological studies.

Of course, the reconstruction algorithms still need to be improved before reaching this application level.
Among the possible improvements, we believe that an interesting extension would be to make our framework \emph{scale-sensitive}.
While some feature point detectors provide a scale space location of the detected feature, we discarded the scale coordinate and used patches of fixed width instead.
This does not depreciate our experiments with FAST points because we used an implementation that is not scale aware, but reconstructions of better quality can probably be achieved by mixing smooth coarse scale patches with finer details.
Additionally, this work did not investigate the issues linked to the geometric transformation invariance enabled by most descriptors.
Our model can be interpreted in terms of reconstruction of canonical image patches that correspond to a reference orientation and scale. 
As far as we have seen, this omission did not create artifacts in our results.
This absence by itself is worth of investigating.

\section*{Acknowledgment}

The \verb+Kata+ image is taken from the iCoseg dataset\footnote{Available online at \url{http://chenlab.ece.cornell.edu/projects/touch-coseg/}} \cite{Batra:2011gz}.
The results obtained with the algorithm proposed in \cite{Weinzaepfel:2011jh} were kindly provided by its authors, and the corresponding original images are courtesy of INRIA through the INRIA Copydays dataset.
The authors would also like to thank Jérôme Gilles from UCLA who contributed greatly to the implementation of the wavelet transforms.
Laurent Jacques is funded by the Belgian Fund for Scientific Research – F.R.S-FNRS.

\ifCLASSOPTIONcaptionsoff
  \newpage
\fi



%

\footnotesize
\bibliographystyle{IEEEtran}
\bibliography{lbd_1bit_reconstruction}

%

\end{document}